\documentclass{article}

\PassOptionsToPackage{numbers, compress}{natbib}

\usepackage[preprint]{neurips_2024}

\usepackage[utf8]{inputenc} %
\usepackage[T1]{fontenc}    %
\usepackage{hyperref}       %
\usepackage{url}            %
\usepackage{booktabs}       %
\usepackage{amsfonts}       %
\usepackage{nicefrac}       %
\usepackage{microtype}      %
\usepackage{xcolor}         %
\usepackage{subcaption}

\usepackage[algo2e,lined,commentsnumbered,ruled]{algorithm2e}

\usepackage{amsmath}
\usepackage{amssymb}
\usepackage{mathtools}
\usepackage{amsthm}
\usepackage{accents}

\usepackage{amsmath,amsfonts,bm}

\def\eqref#1{equation~\ref{#1}}

\def\1{\bm{1}}

\def\vr{{\bm{r}}}

\def\vx{{\bm{x}}}

\def\mA{{\bm{A}}}

\def\mG{{\bm{G}}}

\def\mM{{\bm{M}}}

\def\mX{{\bm{X}}}
\def\mY{{\bm{Y}}}

\DeclareMathAlphabet{\mathsfit}{\encodingdefault}{\sfdefault}{m}{sl}
\SetMathAlphabet{\mathsfit}{bold}{\encodingdefault}{\sfdefault}{bx}{n}

\usepackage[capitalize,noabbrev]{cleveref}
\usepackage{wrapfig}
\theoremstyle{plain}
\newtheorem{theorem}{Theorem}[section]

\theoremstyle{definition}
\newtheorem{definition}[theorem]{Definition}

\theoremstyle{definition}
\newtheorem{property}[theorem]{Property}
\theoremstyle{remark}
\newtheorem*{remark}{Remark}

\usepackage[textsize=tiny]{todonotes}

\newcommand{\pnormal}[2]{\mathcal{N}\left(#1, #2\right)}
\newcommand{\pmnormal}[2]{\mathcal{MN}\left(#1, #2\right)}
\newcommand{\pgamma}[2]{\text{Gamma}\left(#1, #2\right)}
\newcommand{\ppoisson}[1]{\text{Poisson}\left(#1\right)}
\newcommand{\ptpoisson}[2]{\text{TPoisson}_{#1}\left( #2\right)}

\newcommand{\pbernoulli}[1]{\text{Bernoulli}\left(#1\right)}

\newcommand{\pmulti}[2]{\text{Multinomial}_{#1}\left(#2\right)}
\newcommand{\pcat}[2]{\text{Categorical}_{#1}\left(#2\right)}

\newcommand{\pcrt}[2]{\text{CRT}\left(#1, #2\right)}

\newcommand{\taumax}{\tau_{\text{max}}}

\newcommand{\transpose}[1]{{#1}^{\text{T}}}
\newcommand{\diag}[1]{\text{diag}\left(#1\right)}

\newcommand{\rowa}[1]{\renewcommand{\arraystretch}{#1}}

\title{Bayesian Vector AutoRegression with Factorised Granger-Causal Graphs}

\author{%
  He Zhao\thanks{\url{https://hezgit.github.io}} \\
    CSIRO's Data61 \\
  \\
  \And
  Vassili Kitsios \\
   CSIRO’s Environment
  \And
  Terence J. O’Kane
  \\
CSIRO’s Environment
  \And
  Edwin V.~Bonilla
  \\
  CSIRO's Data61
}

\begin{document}

\maketitle

\begin{abstract}
We study the problem of automatically discovering Granger causal relations from observational multivariate time-series data.
Vector autoregressive (VAR) models have been time-tested for this problem, including Bayesian variants and more recent developments using deep neural networks. Most existing VAR methods for Granger causality use sparsity-inducing penalties/priors or post-hoc thresholds to interpret  their coefficients as Granger causal graphs. Instead, we propose a new Bayesian VAR model with a hierarchical factorised prior distribution over binary Granger causal graphs, separately from the VAR coefficients. We develop an efficient algorithm to infer the posterior over binary Granger causal graphs.
Comprehensive experiments on synthetic, semi-synthetic, and climate  data show that our method is more uncertainty aware, has less hyperparameters, and achieves better performance than competing approaches, especially in low-data regimes where there are less observations. 

\end{abstract}
\section{Introduction}
Multivariate time-series (MTS) data consist of  observed samples at multiple timestamps of a set of variables/time-series  and underpins a wide variety of applications in, e.g., economics, healthcare, climatology, and neuroscience. In these applications, it is important to analyse causal structures/relations between the time-series, which not only facilitates a deeper understanding of the data but also provides valuable knowledge for downstream tasks such as prediction and forecasting. In this paper, we are interested in  discovering such causal relations purely from observational MTS data without  conducting costly interventions. For causal discovery on MTS data, Granger Causality (GC)~\cite{granger1969investigating,lutkepohl2005new,shojaie2022granger} is a widely-used tool that  discovers causal relations between variables by quantifying the prediction of the future values of some variables from the past values of the others.
As the fundamental framework for GC, Vector autoregressive (VAR)~\cite{lutkepohl2005new} methods assume that a time series is influenced by the (past) time lags of others and learn autoregression coefficients to capture the influences of the lags. 

Recently, VAR methods based on deep learning~\cite{montalto2015neural,tank2018neural,wang2018estimating,nauta2019causal,khanna2019economy,wu2020discovering,marcinkevivcs2020interpretable,gong2022rhino,fan2023interpretable} have attracted attention by leveraging the flexibility of deep neural networks to capture more complex dynamics of MTS data. 
Despite their impressive results, deep VAR methods may face challenges in practice.
First of all, 
with increased model complexity, deep VARs usually need more data to recover the true GC graphs and may not work well under \textit{low-data regimes} where there are limited observations of variables.
Moreover, as deep VARs often seek for deterministic solutions, they may be less applicable to decision making cases where modelling casual link uncertainty is important.
Finally, causal discovery on MTS data is an unsupervised task without ground-truth graphs, thus, it less clear how to construct a validation set to conduct model selection/hyperparameter tuning, typically necessary in most deep learning methods.

In the above situations,
Bayesian VARs gain unique advantages~\citep{wozniak2016bayesian,miranda2019bayesian}. 
Specifically, being Bayesian is beneficial to capturing epistemic uncertainty in low-data regimes for causal discovery~\cite{annadani2023bayesdag,deleu2023joint} and dealing with interpretability, identifiability issues and in downstream problems such as estimation of causal effects~\cite{geffner2022deep}. 
In addition, Bayesian approaches have principled ways of incorporating prior knowledge by hierarchical prior distributions and reducing the need for model selection or hyperparameter tuning by Bayesian nonparametric techniques ~\cite{orbanz2010bayesian,gershman2012tutorial}.
In this paper, we propose a new Bayesian VAR method for GC that addresses the limitations of deep learning approaches. Our main  innovation is the proposal of a hierarchical Bayesian construction that decomposes the coefficients of a VAR into binary GC graphs indicating the links between variables and real-valued matrices indicating the weights of the links. 
In doing so, our method enjoys several benefits: \textbf{1)} We can capture uncertainty better by modelling the posterior over the binary GC graphs, offering a better way to determining how likely a variable Granger-causes another. 
\textbf{2)} We introduce a factorised Bayesian prior with a principled sparsity-controlling mechanism over binary GC graphs, which is different from  existing approaches that use sparsity penalties~\cite{nicholson2017varx,ahelegbey2016sparse,ghosh2018high,billio2019bayesian} on the coefficients or use post-hoc thresholds to binarise them~\cite{nauta2019causal,marcinkevivcs2020interpretable}.
\textbf{3)} By leveraging Bayesian nonparametric techniques and hierarchical Bayesian models, our method has fewer hyperparameters and facilitates  model selection, thus being more applicable to GC discovery problems where ground-truth graphs are nonexistent.
We conduct extensive experiments on synthetic, semi-synthetic, and climate reanalysis data and compare with various VAR methods for GC. These experiments show that our approach achieves better or comparable performance than others, especially in low-data regime.

\section{Background and Related Work}
In this section, we recap the background of Granger causality with a focus on  Vector AutoRegressive (VAR) models. For more details, we refer the readers to comprehensive surveys such as~\cite{shojaie2022granger,assaad2022survey,gong2023causal}.
Consider a collection of MTS data of $N$ time series or variables in $T$ timestamps, stored in the matrix of $\mX \in \mathbb{R}^{N \times T} = (\vx_1,\dots,\vx_T)$ where $\vx_t \in \mathbb{R}^{N}$ consists of the samples/values of the $N$ variables at timestamp $t\in\{1,\dots,T\}$.
A VAR model for Granger causality~\cite{lutkepohl2005new,hyvarinen2010estimation} assumes that $\vx_t$ can be predicted from the $\taumax$ time lags $\{\vx_{t-1},\dots, \vx_{t-\taumax}\}$ by learning a coefficient matrix $\mA^\tau \in \mathbb{R}^{N \times N}$ for each lag $\tau \in \{1,\dots, \taumax\}$:
\begin{align}
    \label{eq-linvar}
    \vx_t = \sum_{\tau=1}^{\taumax} \mA^\tau \vx_{t-\tau} + \epsilon_t, 
\end{align}
where $\epsilon_t$ is an independent noise variable.
Conventionally, variable $j$ (the parent) does not Granger-cause variable $i$ (the child) ($i,j\in \{1,\dots,N\}$) if and only if for all $\tau$, $A^\tau_{ij}=0$. 
For deterministic VARs,  learning can be done by minimising a regression error: $\min_{\{\mA^\tau\}_{\tau}^{\taumax}} \lVert \vx_t - \sum_{\tau=1}^{\taumax} \mA^\tau \vx_{t-\tau} \rVert^2_2 + \lambda ~ \text{reg}(\{\mA^\tau\}_{\tau}^{\taumax})$ where $\text{reg}(\{\mA^\tau\}_{\tau}^{\taumax})$ is a sparsity-inducing penalty e.g., a group lasso penalty~\cite{yuan2006model,lozano2009grouped}:  $\sum_{i=1,j=1}^N  \lVert A_{i,j}^\tau \rVert_2$. Other alternative penalties can be found at~\cite{nicholson2017varx}.

\textbf{Bayesian VARs} (BVARs) are another important line of research~\cite{breitung2002temporal,george2008bayesian,fox2011bayesian,ahelegbey2016sparse,ghosh2018high,billio2019bayesian,nakajima2013bayesian}, especially in econometrics. For more comprehensive reviews, we refer readers to~\cite{wozniak2016bayesian,miranda2019bayesian}. A standard method may model $\vx_t$ with multivariate normal distributions: 
\begin{align}
\label{eq-bvar}
    \vx_t \sim \pmnormal{\sum_{\tau=1}^{\taumax} \mA^\tau \vx_{t-\tau}}{\Sigma},
\end{align} where various priors can be imposed on $\{\mA^\tau\}_{\tau=1}^{\taumax}$ (e.g., sparsity-inducing priors) and $\Sigma$ (e.g., Inverse-Wishart priors). Learning  BVARs involves inferring the posterior of $\{\mA^\tau\}_{\tau=1}^{\taumax}$ and $\Sigma$. Similar to deterministic VARs, one may need to ``convert'' $\{\mA^\tau\}_{\tau=1}^{\taumax}$ into GC graphs.

\textbf{Deep VARs} have recently become popular, as they use deep neural networks
to model nonlinear dynamics between timestamps~\cite{montalto2015neural,tank2018neural,wang2018estimating,nauta2019causal,khanna2019economy,wu2020discovering,marcinkevivcs2020interpretable,gong2022rhino,fan2023interpretable}. They essentially generalise Eq.~(\ref{eq-linvar}) using:
$x_{it} = f_i\left(\sum_{\tau=1}^{\taumax} \mA^\tau \vx_{t-\tau}\right) + \epsilon_t$, where $f_i$ is a nonlinear function typically implemented by neural networks. 
Different neural network architectures have been explored such as in~\cite{tank2018neural,nauta2019causal,khanna2019economy,marcinkevivcs2020interpretable,bussmann2021neural,fan2023interpretable}.
Despite their promising performance, deep VARs may need a large number of timestamps to discover good causal graphs, require heavy model selection/parameter tuning due to the large number of parameters, and be less uncertainty-aware.
More recently, Bayesian deep VARs have been proposed, such as ACD~\cite{lowe2022amortized}, RHINO~\cite{gong2022rhino}, and Dyn-GFN~\cite{tong2022bayesian}. Although these methods can also be considered as Bayesian approaches, their methodology and focus are quite different from ours. For example, ACD~\cite{lowe2022amortized} focuses on discovering causal
relations across samples with different underlying causal graphs but shared dynamics. RHINO~\cite{gong2022rhino} extends  VAR  by modelling instantaneous causal relations~\cite{runge2020discovering,pamfil2020dynotears} and introducing history-dependent
noise, which we do not consider in this paper. Dyn-GFN~\cite{tong2022bayesian} is a Bayesian approach based on GFlowNets~\cite{bengio2023gflownet} focusing on discovering causal graphs varying with time. There are also recently proposed deep VAR variants focusing on casual discovery on MTS data with missing values~\cite{cheng2023cuts,cheng2024cuts+}, which is beyond the scope of our paper. 

In addition to VARs, there are other  methods for discovering causal structures from MTS data but they are less closely related to ours. To capture instantaneous causal effects that are not modelled by VARs and GC, there are functional casual models such as in~\cite{hyvarinen2010estimation,peters2013causal,pamfil2020dynotears} and methods based on dynamic Bayesian networks (DBNs)~\cite{dean1989model,murphy2002dynamic} or structured VAR models in econometrics~\cite{swanson1997impulse,demiralp2003searching}. For DBNs, we refer readers to surveys such as~\cite{mihajlovic2001dynamic,shiguihara2021dynamic}. Moreover, there are also constraint-based approaches that extend the PC algorithm~\cite{spirtes2000causation} to model time-series data~\cite{runge2018causal,runge2019detecting,runge2020discovering,huang2020causal}.

\section{Method}
The key idea of our methods is introducing a hierarchical Bayesian construction that decomposes the coefficients for each lag of a VAR in Eq.~(\ref{eq-bvar}) into a binary GC graph and a real-valued matrix:
\begin{align}
\label{eq-ours}
        \vx_t \sim \pmnormal{\sum_{\tau=1}^{\taumax} \left(\mA^\tau \odot \mG^\tau\right)  \vx_{t-\tau}}{\Sigma},
\end{align}
where $\mG^\tau \in \{0,1\}^{N \times N}$ is the adjacency matrix for the binary GC graph of lag $\tau$ and $\odot$ denotes the Hadamard product. We further impose the following conjugate prior distributions ~\cite{miranda2019bayesian} on $\mA^\tau$: $\psi^\tau_{i,j} \sim \pgamma{1}{1}, A^{\tau}_{i,j} \sim \pnormal{0}{(\psi^\tau_{i,j})^{-1}}$ and on $\Sigma$: $\lambda_i\sim \pgamma{1}{1}, \Sigma=\diag{\lambda_1, \dots, \lambda_N}^{-1}$ where $\diag{\lambda_1, \dots, \lambda_N}$ returns a matrix with its diagonal elements as $\lambda_1, \dots, \lambda_N$.

We highlight $\mathbf{A} \odot \mathbf{G}$ as the key proposal of our model, where
we decouple the impact of variable $j$ on $i$ into two components: $G^\tau_{i,j} \in \{0,1\}$ indicating whether there is a link between $i$ and $j$ and $A^\tau_{i,j} \in \mathbb{R}$ indicating the weight of the link. If $G^\tau_{i,j}=0$, $j$ does not impact $i$ in lag $\tau$ regardless of the value of $A^\tau_{i,j}$ while if $G^\tau_{i,j}=1$, $A^\tau_{i,j}$ captures the influence from $j$ to $i$. 
We also note that variable $j$ does not Granger-cause variable $i$ ($i,j\in \{1,\dots,N\}$) if and only if for all $\tau$, $G^\tau_{i,j}=0$.
The proposed decoupling leads to several appealing benefits. 
\textbf{1)} Now $\mathbf{A} \odot \mathbf{G}$ contains the coefficients of our VAR and it is intrinsically sparse as $\mathbf{G}$ is binary. Thus, no sparsity-inducing penalties~\cite{nicholson2017varx,ahelegbey2016sparse,ghosh2018high,billio2019bayesian} or post-hoc heuristics~\cite{nauta2019causal,marcinkevivcs2020interpretable} are needed.
\textbf{2)} It gives us the flexibility of modelling $\mathbf{A}$ and $\mathbf{G}$ with different models. This flexibility enables us to model $\mathbf{G}$ with an informative hierarchical prior taking a factorised form (as shown later) providing informative prior knowledge which is useful in low-data regimes.
\textbf{3)} Modelling $\mathbf{G}$ with Bernoulli distributions directly enables us to learn the probability of a casual link between two variables in a principled way. In our method, one can easily quantify the confidence interval and standard deviations of a link.

Despite the above benefits, learning binary graphs is a nontrivial problem. Given a binary GC graph with $N$ variables, a naive approach might need to search from a space of $\mathcal{O}(2^{N^2})$ possible solutions. Therefore, it is important to use a suitable Bayesian prior $p(\{\mG^\tau\}^{\taumax}_\tau)$ that can lead to an efficient inference algorithm of $p(\{\mG^\tau\}^{\taumax}_\tau| \mX)$.

\subsection{Poisson Factorised Granger-Causal Graph Model}
Now we introduce our Bayesian construction on GC graphs that can be integrated into VARs.
To assist clarity, we discuss our method with only one lag, i.e., $\taumax=1$, temporally omitting the notation of lag $\tau$, and introduce the extension to multiple lags later.
The general idea is that we assume a binary GC graph $\mG \in \{0,1\}^{N \times N}$ is a sample of a probabilistic factorisation model with $K$ latent factors: $\mG \sim p(\Theta \transpose{\Phi})$ where $\Theta \in \mathbb{R}_+^{N \times K}$  each entry of which $\theta_{i,k}$ indicates the weight of the $k^{\text{th}}$ factor for variable $i$ of being a child in a GC relation and $\Phi \in \mathbb{R}_+^{N \times K}$ each entry of which $\phi_{j,k}$ indicates the weight of the $k^{\text{th}}$ factor for variable $j$ of being a parent in a GC relation. In this way, whether $j$ Granger-causes $i$ depends on their interactions with all the $K$ factors: $G_{i,j} \sim p\left(\sum_{k=1}^K \theta_{i,k} \phi_{j,k}\right)$. Conditioned on $\Theta$ and $\Phi$, we have: $p(\mG | \Theta, \Phi) = \prod_{i=1}^N \prod_{j=1}^N p(G_{i,j} | \Theta, \Phi)$, meaning that the links in $\mG$ can be generated independently.

As $\mG$ is discrete, it is natural to leverage the idea of Poisson factor analysis (PFA), a widely-used factorisation approach for discrete data~\cite{canny2004gap,zhou2012beta,gopalan2014content} with a rich set of statistical tools in probabilistic modelling and inference. As the original PFA models count-valued data with the Poisson distribution, it would be a model misspecification if this is used for binary data in our case.
To address this issue, inspired by~\cite{zhou2015infinite}, we introduce a new link function named Generalised Bernoulli Poisson Link (GBPL) that thresholds a random Poisson variable $m$ at $V \in \{1,2,\dots\}$ to obtain a binary variable $b$.
\begin{definition}{(Generalised Bernoulli Poisson Link)}
\begin{align}
    m \sim \ppoisson{\gamma},
    b = \mathbf{1}(m \ge V), \nonumber
\end{align} where $\mathbf{1}(\cdot)$ is function returning one if the condition is true otherwise zero. 
\end{definition}

\begin{property}
\label{pro-marignal}
Given $\gamma$ and $V$, one can marginalise $m$ out to get:
$b \sim \pbernoulli{1 - \sum_{v=0}^{V-1} \frac{e^{-\gamma} \gamma^v}{v!}}$.
\end{property}
\begin{remark}
As $b=0$ if and only if $m < V$, $p(b=0) = \sum_{v=0}^{V-1} p(m=v)$, thus, $p(b=1) = 1 - \sum_{v=0}^{V-1} p(m=v)$. Moreover, as $\mathbb{E}(b)=p(b=1)$, larger $V$ leads to lower expected probability of $b$ being one, under the same $\gamma$.
\end{remark}

\begin{property}
\label{pro-cond}
Given $b$, the conditional posterior of $m$ is in close-form:
\begin{align}
m \sim 
\begin{dcases}
\ptpoisson{V}{\gamma}, & \text{if } b>0\\
\pcat{V}{\left[\dots, f(v, \gamma),\dots\right]},              & \text{otherwise}\nonumber
\end{dcases}
\end{align}
where $\ptpoisson{V}{\gamma}$ is the Poisson distribution with parameter $\gamma$ left-truncated at $V$ (i.e., the samples from that Poisson distribution are greater than or equal to $V$) and $f(v, \gamma) = \frac{\frac{e^{-\lambda} \lambda^v}{v!}}{\sum_{v'=0}^{V-1} \frac{e^{-\lambda} \lambda^{v'}}{v'!}}$ is the normalised Poisson probability mass function.
\end{property}
\begin{remark}
If $b>0$, $m \ge V$ almost surely (a.s.) and one can sample $m$ from the truncated Poisson distribution at $V$ efficiently by computing the inverse Poisson cumulative distribution function~\cite{giles2016algorithm}. If $b=0$, $m \in \{0,\dots, V-1\}$ is sampled from the categorical distribution with normalised Poisson probability masses. The close-form conditional posterior contributes to the development of an efficient algorithm of our model.
\end{remark}

\begin{property}
When $V$ is set to 1, GBPL reduces to the link function proposed in~\cite{zhou2015infinite}.
\end{property}
\begin{remark}
Compared with the link function in~\cite{zhou2015infinite}, GBPL is a more flexible method with an additional parameter $V$. As shown later, $V$ controls the sparsity of generated graphs, which is essential for our problem. 
\end{remark}

With the help of GBPL, we propose to impose the following hierarchical Bayesian prior $p(\mG)$:
\begin{align}
        r_k \sim \pgamma{1/K}{1/c}~~,
        \label{eq-base-dis}
    \theta_{i,k} \sim \pgamma{a_i}{1/d_k},~~\phi_{j,k} \sim \pgamma{b_j}{1/e_k}, \\
    \label{eq-pmf}
    M_{i,j} \sim \ppoisson{\sum_{k=1}^K r_k\theta_{i,k}\phi_{j,k}},~~
    G_{i, j} = \mathbf{1}(M_{i, j} \ge V),
\end{align}
where noninformative gamma priors $\pgamma{1}{1}$ are used for $a_i$, $b_j$, $d_k$, $e_k$, and $c$.

In the above model, 
a new variable $r_k$ is introduced to capture the global popularity of  the $k^\text{th}$ factor~\cite{yang2012community,yang2014structure,zhou2015infinite}. Mathematically, the construction on $\Theta$, $\Phi$, and $\vr$ can be viewed as the truncated version of a gamma process~\citep{ferguson1973bayesian,wolpert2011stochastic,zhou2015infinite} on a product space $\mathbb{R}_+ \times \Omega$: $\mathfrak{G} \sim \Gamma\text{P}(\mathfrak{G}_{a, b}, 1/c)$,  
where $\Omega$ is a complete separable metric space, $c$ is the concentration parameter, $\mathfrak{G}_{a,b}$ is a finite and continuous base measure over $\Omega$.
The corresponding L{\'e}vy measure is  $\nu(drd\theta d\phi) = r^{-1} e^{-c r}dr \mathfrak{G}_{a,b}(d\theta d\phi)$. 
In our case, a draw from the $\mathfrak{G}_{a,b}$ is a pair of $\theta_{:,k}$ and $\phi_{:,k}$ as shown in Eq.~(\ref{eq-base-dis}) where $\theta_{:,k} = [\theta_{1,k}, \dots, \theta_{N, k}]$
$\theta_{:,k} = [\theta_{1,k}, \dots, \theta_{N, k}]$ and $\phi_{:,k} = [\phi_{1,k}, \dots, \phi_{N, k}]$.
A draw from the gamma process is a discrete distribution with countably infinite atoms from the base measure: $\mathfrak{G} = \sum_{k=1}^\infty r_k \delta_{\theta_{:,k}, \phi_{:,k}}$ and $r_k$ is the weight of the $k^{\text{th}}$ atom. Although there are infinite atoms, the number of atoms with $r_k$ greater than $\rho \in \mathbb{R}_+$ follows $\text{Poisson}(\int_{\rho}^{\infty} r^{-1} e^{-cr} \text{d}r)$ and the expectation of Poisson decreases when $\rho$ increases. In other words, the number of atoms that have relatively large weights will be finite and small, thus, a gamma process based model has an inherent shrinkage
mechanism. In our case, if we set the maximum number of latent factors $K$ (i.e., the truncation level) large enough, the model will automatically learn the number of active factors. 

Given the specification of the prior distributions, one can see that $\mathbb{E}\left(\sum_{k=1}^K r_k\theta_{i,k}\phi_{j,k}\right) = 1$ as $\mathbb{E}(\theta_{i,k}) = \mathbb{E}(\phi_{j,k}) = 1$ and $\mathbb{E}(r_k) = 1/K$. Therefore, according to Property~\ref{pro-marignal} of GBPL, the expected sparsity of $\mG$ in the prior distribution is $N^2 \left(1 - \sum_{v=0}^{V-1} \frac{e^{-1}}{v!}\right)$. Note that $V \in \{1,2,\dots\}$ is a natural number that controls the sparsity, i.e., larger value of $V$ leads to sparser graphs. Theoretically, $V$ can be arbitrarily large, however, large $V$ makes the sampling of the truncated Poisson distribution (Property~\ref{pro-cond}) harder. We empirically find that when $V>3$, the sampled GC graphs from the posterior nearly have zero links, thus, we set $V \in \{1,2,3\}$. Note that $V$ is a parameter that only takes bounded discrete values, while the regularisation weights in other VARs are usually continuous parameters taking infinite values.

Finally, we refer to the model in Eq.~(\ref{eq-ours}) and~(\ref{eq-pmf}) as Poisson Factorised Granger-Causal Graph (PFGCG) and denote $\mG \sim \text{PFGCG}(V)$. In the case of multiple lags, we summarise the model as: 
\begin{align}
\label{eq-ours-2}
        \mG^\tau \sim \text{PFGCG}^\tau(V),~~
        \vx_t \sim \pmnormal{\sum_{\tau=1}^{\taumax} \left(\mA^\tau \odot \mG^\tau\right)  \vx_{t-\tau}}{\Sigma},
\end{align}
where we have a separate generative process for the GC graph at each lag $\tau$.

\subsection{Inference via Gibbs Sampling}
\label{sec-inference-short}
Here we introduce how to learn PFGCG by Bayesian inference via Gibbs sampling. Recall that Gibbs sampling~\cite{casella1992explaining} is a widely-used Markov chain Monte Carlo (MCMC) method that takes the samples of a variable from its conditional posteriors conditioned other variables, thus, requiring ``easy'' sampling from the conditional posteriors. When it converges, the samples from a Gibbs sampler will be equivalent to those sampled from the true posterior. In our model, with the help of several augmentation techniques between Poisson and gamma distributions~\cite{zhou2012beta,zhou2015infinite}, PFGCG enjoys local conjugacy and all the variables have close-form conditional posteriors, which facilitate an efficient inference algorithm. Here we highlight the sampling of $\mG^\tau$ and leave the other details in Section~\ref{sec-inference}.

An entry $G^\tau_{i,j}$ in $\mG^\tau$ is involved in the generative process of data as in Eq.~(\ref{eq-bvar}) and has a Bernoulli prior according to Eq.~(\ref{eq-pmf}). Therefore, by denoting $p(G^\tau_{i,j}=0|-)=s^{\tau,0}_{i,j}$ and $p(G^\tau_{i,j}=1|-)=s^{\tau,1}_{i,j}$ ($-$ stands for all the other variables), we can derive:
\begin{align}
    \label{eq-sample-g}
    s^{\tau,0}_{i,j} = \sum_{v=0}^{V-1} \frac{e^{-q^\tau_{i,j}} (q^\tau_{i,j})^v}{v!}, ~~
    s^{\tau,1}_{i,j} = e^{-\frac{1}{2} \left((A^\tau_{i,j})^2  \lambda_i U^\tau_j - 2A^\tau_{i,j}\lambda_i W^{\tau}_{i,j} \right)} \left(1 - s^{\tau,0}_{i,j}\right), 
\end{align}
where $q^\tau_{i,j} = \sum_{k=1}^K \theta^\tau_{i,k} r^\tau_k \phi^\tau_{j,k}$, $U^\tau_j = \sum_{t=1}^T x^2_{j, t-\tau}$, $W^{\tau}_{i,j}=\sum_{t=1}^T x^{\neg \tau, \neg j}_{i,t} x_{j, t-\tau}$, $x^{\neg \tau, \neg j}_{i,t} = x_{i,t} - \sum_{j' \neq j}^N A^\tau_{i,j'} G^{\tau}_{i,j'} x_{j', t-\tau} - \sum_{\tau' \neq \tau}^{\taumax} \sum_{j'=1}^{N} A^{\tau'}_{i,j'} G^{\tau'}_{i,j'} x_{j', t-\tau'}$. We can then sample $G^\tau_{i,j} \sim \pbernoulli{s^{\tau,1}_{i,j} /\left(s^{\tau,0}_{i,j}+ s^{\tau,1}_{i,j}\right)}$.
With the above conditional posterior, one can sample the entries of $\mG^{\tau}$ one by one
using Eq.~(\ref{eq-sample-g}). After each sample, we only need to update $W^{\tau}_{i,j}$ and the other statistics can be updated after all the entries are sampled. Therefore, in one Gibbs sampling iteration, the complexity of sampling $\mG^{\tau}$ is $\mathcal{O}(N^2)$. As shown in Algorithm~\ref{alg}, the whole complexity of each Gibbs sampling iteration of our model is $\mathcal{O}(N^2(V+K)\taumax + T\taumax^2)$, where $N$, $T$, $V$, $K$, $\taumax$ are the number of variables, the number of timestamps, the truncation level (a hyperparameter of ours), the number of factors, and the maximum number of lags, respectively.

\section{Experiments}

\subsection{Experimental Settings}
\label{sec-settings}

\paragraph{Settings of Our Method}
We use 10,000 as the maximum Gibbs sampling iterations where the first 5,000 are burn-in iterations and we then collect the samples from the  conditional posteriors of the graphs in every 10 iterations\footnote{As shown in Section~\ref{sec-complexity}, our method converges in much less iterations.}, which are stored in $\mY \in \mathbb{R}_+^{N \times N \times \taumax \times H}$ ($H=500$ is the number of collections).
The Bernoulli conditional posterior probability of a link between $i$ and $j$ at lag $tau$ in collection $h \in \{1,\dots,H\}$ is computed by Eq.~(\ref{eq-sample-g}) as:  $Y[i,j,\tau,h] = \frac{s^{\tau,1}_{i,j}}{s^{\tau,0}_{i,j}+ s^{\tau,1}_{i,j}}$.
Given the collections, to compare with other methods, we compute the averaged probability of the discovered GC graph by $\text{mean}(\text{max}(\mY, \text{dim}=`\tau'), \text{dim}=`h')$~\cite{marcinkevivcs2020interpretable}.
As PFGCG has an intrinsic shrinkage mechanism on $K$, we set $K=50$ that is empirically large enough for our experiments. The only hyperparameter that we need to tune is $V$, which we vary in $\{1,2,3\}$.
\paragraph{Baselines}
As ours is a VAR approach for GC, we mainly include baselines that are also based on the VAR framework in our comparison.
\textbf{1)} We compare with the widely-used VAR with F-tests for Granger causality and the Benjamini-Hochberg procedure~\cite{benjamini1995controlling} for controlling the false discovery rate (FDR) (at $q=0.05$) denoted as VAR (FBH) and implemented in the statsmodels library~\cite{seaboldeconometric}.
\textbf{2)} For Bayesian methods, we compare with two classic approaches based on Eq.~(\ref{eq-bvar}) but with different prior distributions:
BVAR with diffuse/noninformative priors on the coefficients, named BVAR(d), i.e.,  $(\{\mA^\tau\}_{\tau}^{\taumax}, \Sigma) \propto |\Sigma|^{-\frac{N + 1}{2}}$ in Eq.~(\ref{eq-bvar})~\cite{litterman1986forecasting,miranda2019bayesian}, whose posterior has an analytical form.
BVAR with conjugate priors on the coefficients, named BVAR(c): $\psi^\tau_{i,j} \sim \pgamma{1}{1}, A^{\tau}_{i,j} \sim \pnormal{0}{(\psi^\tau_{i,j})^{-1}}$ and for $\Sigma$: $\lambda_i\sim \pgamma{1}{1}, \Sigma=\diag{\lambda_1, \dots, \lambda_N}^{-1}$. This is equivalent to an ablation of our model without $\{\mG^\tau\}_{\tau}^{\taumax}$, for which we use the Gibbs sampling with the same settings as ours.
For deep VARs, we compare with a method with component-wise statistical recurrent units (SRU)~\cite{oliva2017statistical} and its improved version (economy SRU, eSRU)~\cite{khanna2019economy} with sample-efficient architectures. The important hyperparameters of SRU and eSUR are the strengths ($\mu_1$, $\mu_2$, $\mu_3$) of three regularisation terms.
We also compare with GVAR~\cite{marcinkevivcs2020interpretable} that uses self-explaining neural networks~\cite{alvarez2018towards} and converts the weights in the neural networks into binary GC graphs with a heuristic stability-based procedure. GVAR is reported to have state-of-the-art performance.
GVAR has two important regularisation hyperparameters $\lambda$ and $\gamma$. For SRU, eSUR, and GVAR, the original implementations are used. For the baselines, we either use their original settings or follow these in~\cite{marcinkevivcs2020interpretable}, shown  in Table~\ref{tb-setting} of the appendix. For non-VAR methods, the most recent one is PCMCI$^+$~\cite{runge2020discovering}, which does not outperform eSRU in the experiments of~\cite{gong2022rhino}, thus, is not included in our comparison.

\paragraph{Evaluation Metrics}
Following~\cite{khanna2019economy,marcinkevivcs2020interpretable} that aggregate graphs at multiple lags into one,
we use four metrics to compare the discovered GC graph of a method on a dataset with the ground-truth graph. For all the baseline methods, we compute the score of a discovered GC graph from their learned VAR coefficients. For our method, the score of a GC graph is the mean of the Bernoulli posterior. We report the areas under receiver operating characteristic (AUROC) and precision-recall (AUPRC) curves by comparing  the score of a discovered GC graph to the ground-truth graph. Moreover, as mentioned before, VAR (FBH) and GVAR use specific post-hoc processes to convert coefficients to binary GC graphs, thus, we also report the structural Hamming distance (SHD) between the discovered binary GC graph and the ground-truth one. To compute SHD for ours, we sample a graph from the Bernoulli posterior mean. Note that unlike AUCROC and AUPRC, SHD is biased to the sparsity of the ground-truth graph, e.g., for a sparse ground-truth graph, a method always predicting no links achieves low SHD.  To measure the predictive uncertainty of the Granger-casual graphs discovered by different approaches, we report the calibration error (CE)~\cite{guo2017calibration}, which has been a widely used metric for model uncertainly and confidence~\cite{liu2020simple}. CE examines the difference between the model’s probability and the true probability given the model’s output, whose definition is shown in Definition 2.1 of~\cite{kumar2019verified}. We consider the casual discovery task with $N$ variables as a binary classification problem with $N^2$ samples, i.e., predicting a Granger-causal link between a pair of variables and then compute CE accordingly. For AUCROC and AUPRC, higher values indicate better performance and for SHD and CE, lower values are better.

\paragraph{Model Selection and Parameter Tuning}
To conduct model selection for each method, we split the input MTS data into a training set (first 80\% timestamps) and a test set (remaining 20\%)~\cite{gong2022rhino}. We train a method on the training set and use the learned model to conduct one-step prediction on the test set. We then use the mean square error (MSE) on the test set and select the parameters of a method that give the best MSE. Our model selection is different from that of GVAR~\cite{marcinkevivcs2020interpretable}, where the best model is selected by comparing with the ground-truth graphs and report the best achievable performance. We report best MSE in our experiments as a reference but do not view it as the main metrics as our primary goal is not on predictions.
The parameter space where we search for each method is shown in Table~\ref{tb-setting} of the appendix.

\subsection{Synthetic Data}

We conduct experiments on the synthetic data generated from a VAR model specified in Eq.~(\ref{eq-linvar}) to test whether our method is able to discover the ground-truth graphs. Given $N=16$ and $\taumax=6$, we  construct $\{\mA^\tau\}_{\tau=1}^{\taumax}$ by first specifying the nonzero entries (i.e, the ground-truth casual graphs) and then for each nonzero entry we sample $A^\tau_{i,j} \sim \text{uniform}(0.1, 0.2)$. 
We generate $T=1,000$ samples accordingly, initialise $\vx_0$ from a standard normal distribution and sample $\epsilon_t \sim \pnormal{0}{0.01}$. 
We show the results of our method in Figure~\ref{fig-syn-1} in the appendix, where we also fit a randomly initialised VAR~\cite{seaboldeconometric} to the data as a reference. It can be seen that the ground-truth graphs at different lags have diverse patterns and our method discovers them well (also reflected by better AUCROC and AUPRC). Unlike VAR, our method directly discovers binary graphs without using thresholds or tests.

\subsection{Semi-synthetic Benchmark Datasets}

\begin{figure*}[t]
        \centering
         \begin{subfigure}[b]{0.48\linewidth}
                 \centering
                 \includegraphics[width=0.99\textwidth]{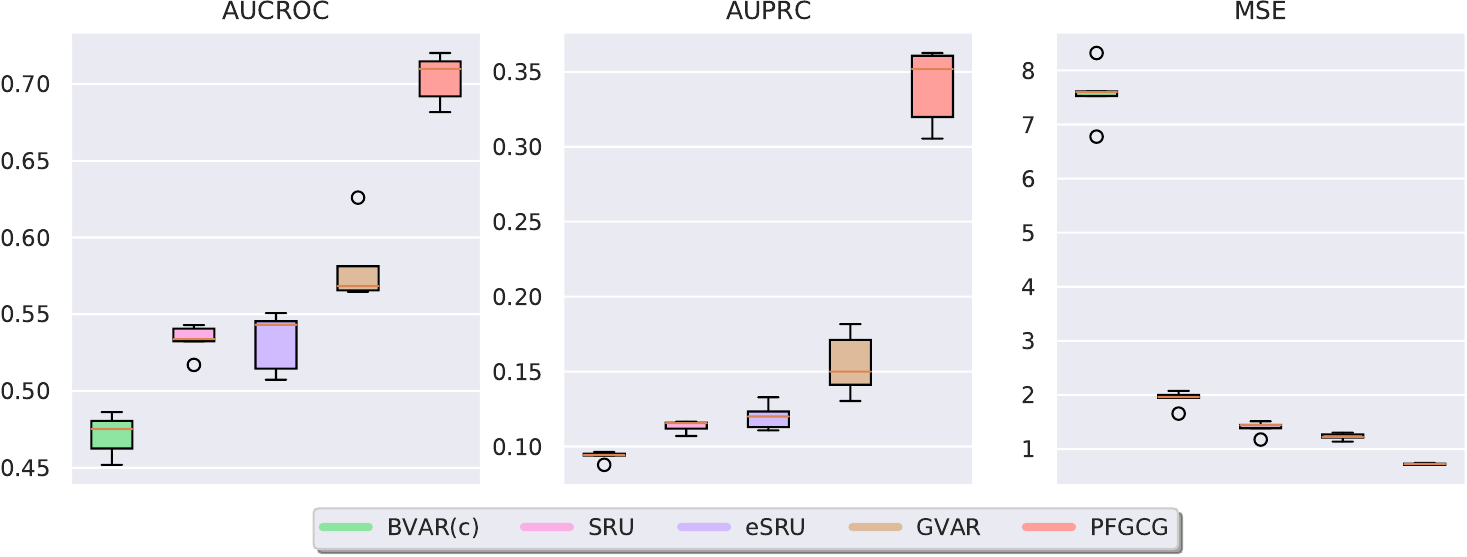}
                                           \caption{Lorenz 96 $T=100$}

         \end{subfigure}
         \hspace{-0.5cm}
         \begin{subfigure}[b]{0.53\linewidth}
                 \centering

                 \includegraphics[width=0.9\textwidth]{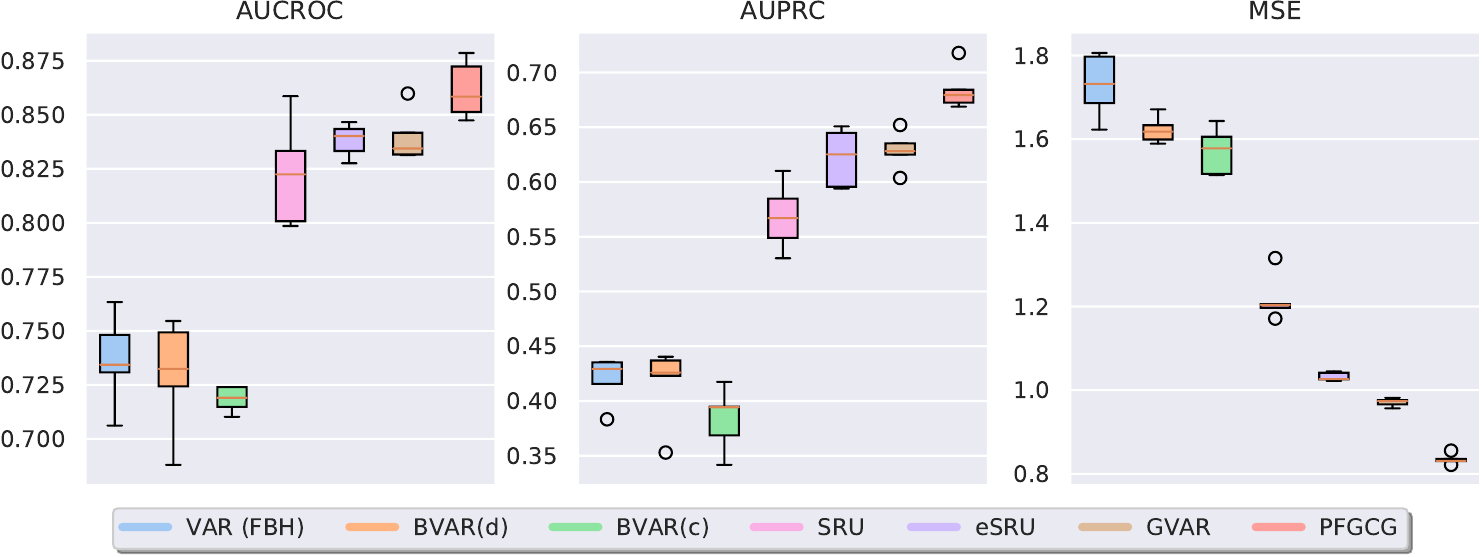}
                                   \caption{Lorenz 96 $T=500$}

         \end{subfigure} \\
         \begin{subfigure}[b]{0.48\linewidth}
                 \centering
                 \includegraphics[width=0.99\textwidth]{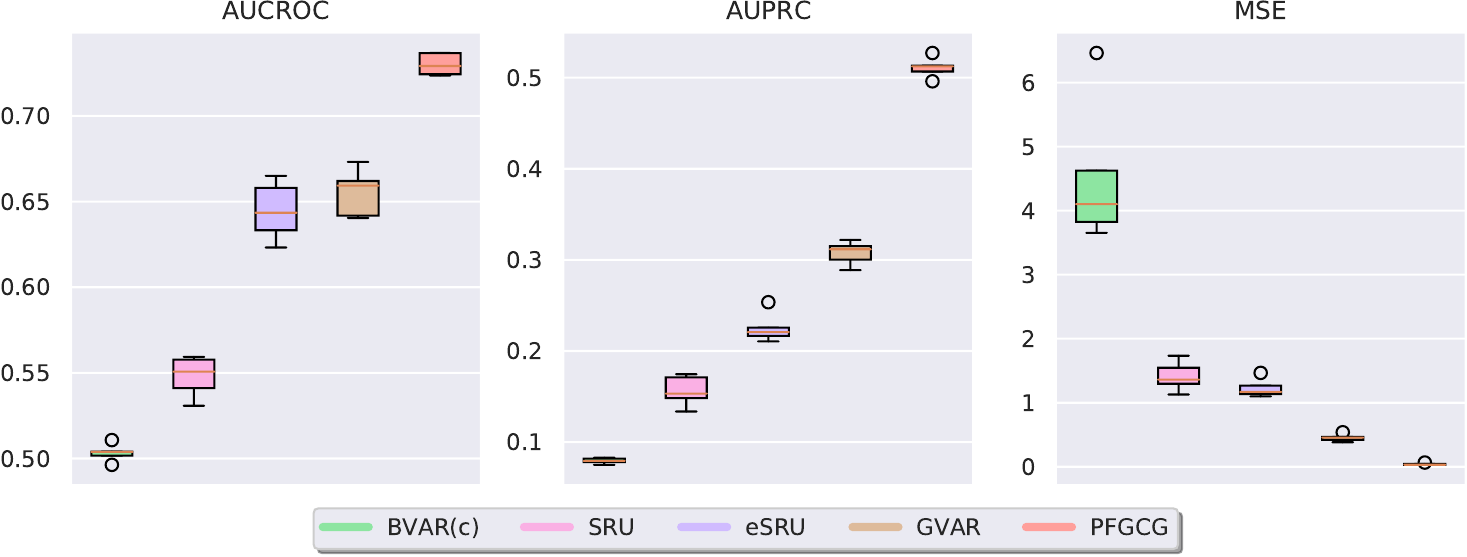}
                                  \caption{Lotka–Volterra $T=200$}

         \end{subfigure}
         \hspace{-0.5cm}
         \begin{subfigure}[b]{0.53\linewidth}
                 \centering
                 \includegraphics[width=0.9\textwidth]{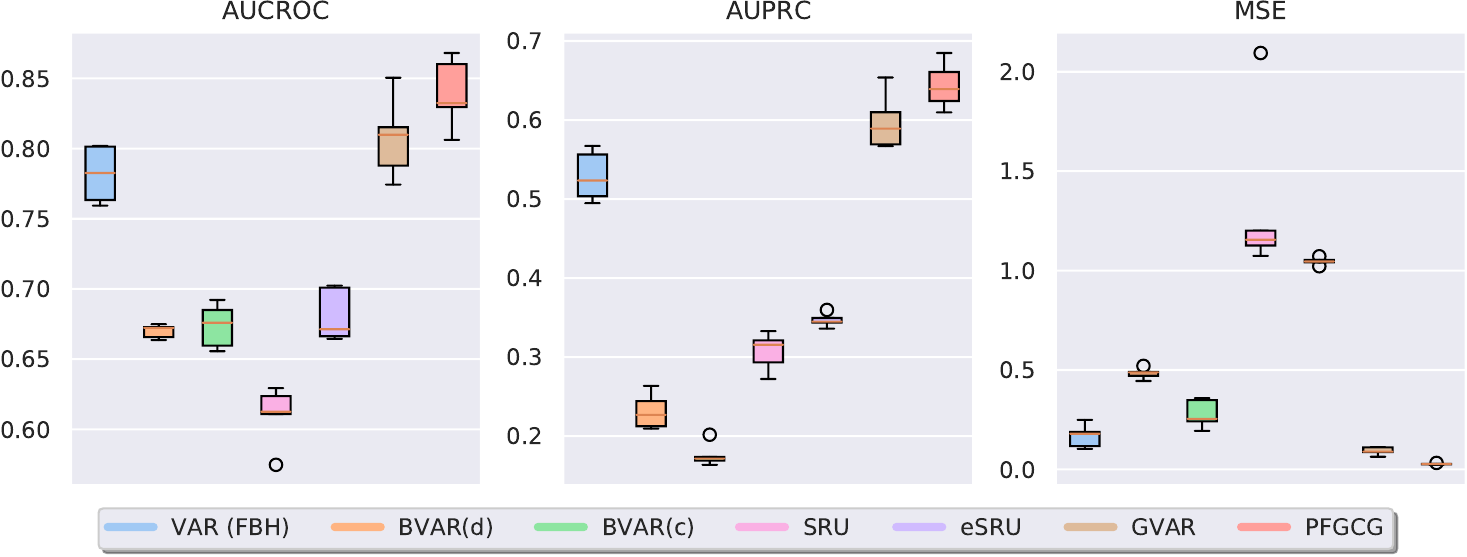}
                                   \caption{Lotka–Volterra $T=500$}

         \end{subfigure} 
\caption{Results on Lorenz 96 and Lotka–Volterra. VAR (FBH) and BVAR(d) failed to learn when $T=100$ on Lorenz 96 and $T=200$ on Lotka–Volterra.}
\label{fig-l96}
 \vspace{-0.5cm}
\end{figure*}

\begin{wrapfigure}{r}{0.5\textwidth}
\captionsetup[subfigure]{justification=centering}
        \centering
         \begin{subfigure}[b]{0.99\linewidth}
                 \centering
                 \includegraphics[width=0.99\textwidth]{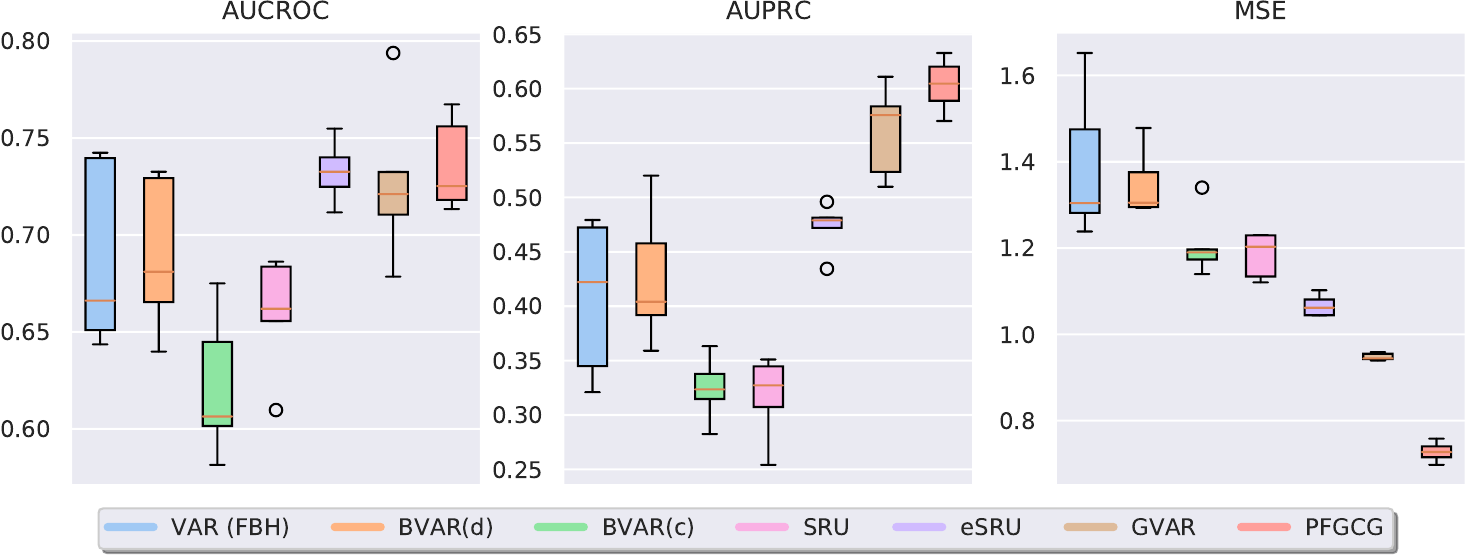}
         \end{subfigure}
\caption{Results on FMRI.}
\label{fig-fmri}
 \vspace{-0.2cm}
\end{wrapfigure}

For more comprehensive quantitative comparisons, we conduct our experiments on three widely-used benchmark datasets. For each dataset we use 5 replicates and we vary $\taumax={1,3,5}$ for all the methods. \textbf{1)} Lorenz 96~\cite{lorenz1996predictability} is a standard benchmark synthetic MTS dataset for GC, which is generated from the following nonlinear differential equations: $\frac{d x_{i,t}}{dt} = (x_{i+1,t} - x_{i-2,t}) x_{i-1,t} - x_{i,t} +F, \text{~for~} 1 \le i \le N,$
where $F$ is a constant that models the magnitude of the external forcing. The system
dynamics become increasingly chaotic for higher values of $F$~\cite{karimi2010extensive}.
Following~\cite{marcinkevivcs2020interpretable}, we set $F=40$, which is a more difficult case. 
We use $T=\{100,500\}$ where $T=100$ is to mimic low-data regimes. 
\textbf{2)} Following~\cite{marcinkevivcs2020interpretable}, we evaluate the methods on another synthetic dataset generated by the Lotka–Volterra model~\cite{bacaer2011lotka}, where we use $N=40$ and $T=\{200,500\}$. For the other parameters of the Lotka–Volterra model, we use the same settings as in~\cite{marcinkevivcs2020interpretable}.
\textbf{3)}
We consider the FMRI dataset with realistic simulations of blood-oxygen-level dependent (BOLD) time series~\cite{smith2011network}. These were generated using the dynamic causal modelling functional magnetic resonance imaging (fMRI) forward model. Following~\cite{khanna2019economy,marcinkevivcs2020interpretable}, we use 5 replicates from the simulation no. 3 of the original dataset, where $N=15$ and $T=200$ are pre-specified. We notice that \cite{cheng2024causaltime} recently introduces a few new benchmark datasets with many timestamps (e.g, from 8,000 to 50,000). Our focus is on low-data regimes with less timestamps and their ground-truth graphs are undirected (i.e., the adjacency matrices are symmetric) while our method discovers directed graphs, therefore, these datasets are less applicable to our problem.

\begin{table*}[]
\rowa{1.3}
\centering
\caption{SHD and CE. Means and standard derivations are computed over 5 replicates on each dataset.}
\label{tb-shd}
\resizebox{0.85\linewidth}{!}{
\begin{tabular}{@{}ccccccc@{}} \toprule

          & \multicolumn{2}{c}{L96}                                  & \multicolumn{2}{c}{LV}                                  & FMRI            \\ 
          & \multicolumn{1}{c}{$T=100$}           & $T=500$          & \multicolumn{1}{c}{$T=200$}           & $T=500$         &                 \\ \midrule
& \multicolumn{5}{c}{SHD$\downarrow$} &\\ \midrule
VAR (FBH) & \multicolumn{1}{c}{-}                 & 98.40$\pm$2.42   & \multicolumn{1}{c}{-}                 & 74.20$\pm$10.46 & 28.80$\pm$1.33  \\ 
GVAR      & \multicolumn{1}{c}{389.60$\pm$220.64} & 127.40$\pm$76.84 & \multicolumn{1}{c}{279.00$\pm$104.13} & 82.80$\pm$24.10 & 71.60$\pm$21.82 \\ 
PFGCG     & \multicolumn{1}{c}{117.76$\pm$3.34}   & \textbf{71.83$\pm$4.07}   & \multicolumn{1}{c}{\textbf{67.01$\pm$2.46}}    & \textbf{45.05$\pm$4.40}  & \textbf{24.26$\pm$0.76}  \\
\midrule
& \multicolumn{5}{c}{CE$\downarrow$} &\\ \midrule
BVAR (d) &       -         & 0.10$\pm$0.01  & - & 0.11$\pm$0.01 & 0.11$\pm$0.02 \\ 
GVAR      & 0.07$\pm$0.01  & 0.15$\pm$0.01  & 0.08$\pm$0.01 & 0.10$\pm$0.01 & 0.19$\pm$0.01 \\ 
PFGCG     & \textbf{0.11$\pm$0.01}  &  \textbf{0.08$\pm$0.01}  &  \textbf{0.05$\pm$0.01} &  \textbf{0.04$\pm$0.01} &  \textbf{0.07$\pm$0.01} 
\\
\bottomrule
\end{tabular}
}
\vspace{-0.3cm}
\end{table*}

We show the results of AUCROC and AUPRC in Figure~\ref{fig-l96}, and~\ref{fig-fmri}. where we report $\taumax=5$ in the main paper and the results of $\taumax=1, 3$ are in the appendix. The results of SHD and CE are shown in Table~\ref{tb-shd}. For the SHD results, we show the methods whose coefficients can be converted to sparse graphs and for the CE results, we focus on the best-performing methods in terms of AUCROC and AUPRC. In general, we can see that our PFGCG performs the best on all the datasets in terms of the four metrics. Our method's performance advantage is more significant when there are less timestamps (e.g., $T=100$ on Lorenz 96 and $T=200$ on Lotka–Volterra), demonstrating the appealing properties of using the proposed Bayesian model in low-data regimes. Although the mean square error (MSE) is not the focus here, our method general achieves the lowest achievable MSE as well\footnote{We do model selection for all the methods according to their MSE on the test set. Therefore, MSE reported here is just a reference.}. In the comparison between PFGCG and BVAR(c) that is equivalent to PFGCG without the binary GC graphs, one can observe that PFGCG consistently outperforms BVAR(c), showing the effectiveness of introducing binary GC graphs into the model.
For CE, it can be seen that our method achieves the lowest CE on all the datasets. Although the AUCROC/AUPRC scores of  BVAR (d) are no better than GVAR's, it outperforms GVAR in terms of CE, demonstrating that Bayesian methods have intrinsic advantages on uncertainty.

\begin{wrapfigure}{r}{0.6\textwidth}
\captionsetup[subfigure]{justification=centering}
        \centering
         \begin{subfigure}[b]{0.32\linewidth}
                 \centering
                 \includegraphics[width=0.99\textwidth]{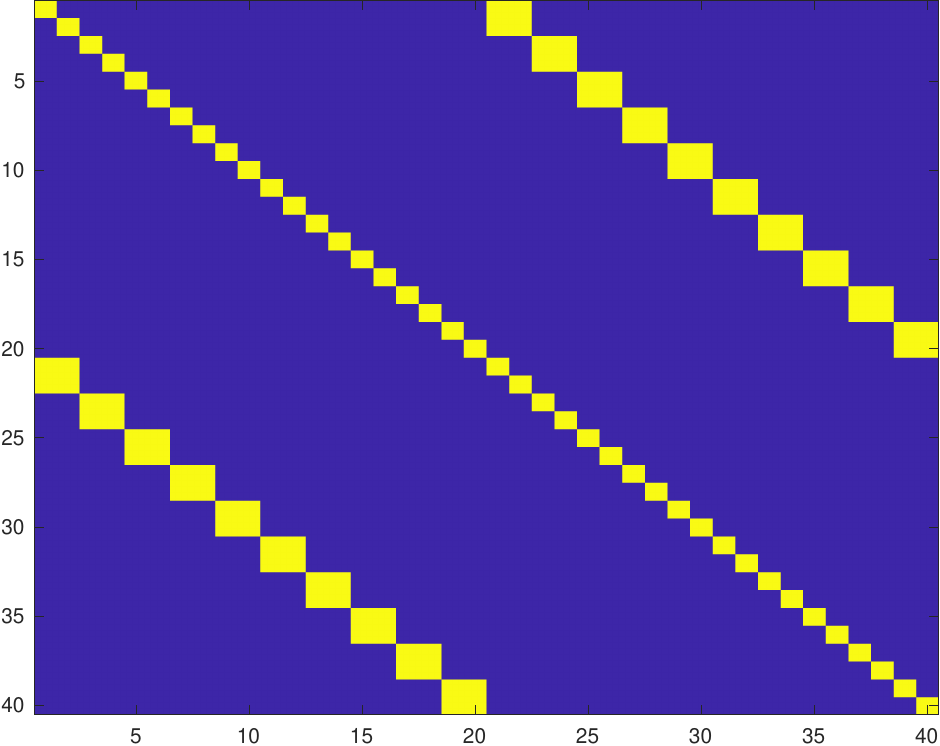}
         \end{subfigure}
         \begin{subfigure}[b]{0.32\linewidth}
                 \centering
                 \includegraphics[width=0.99\textwidth]{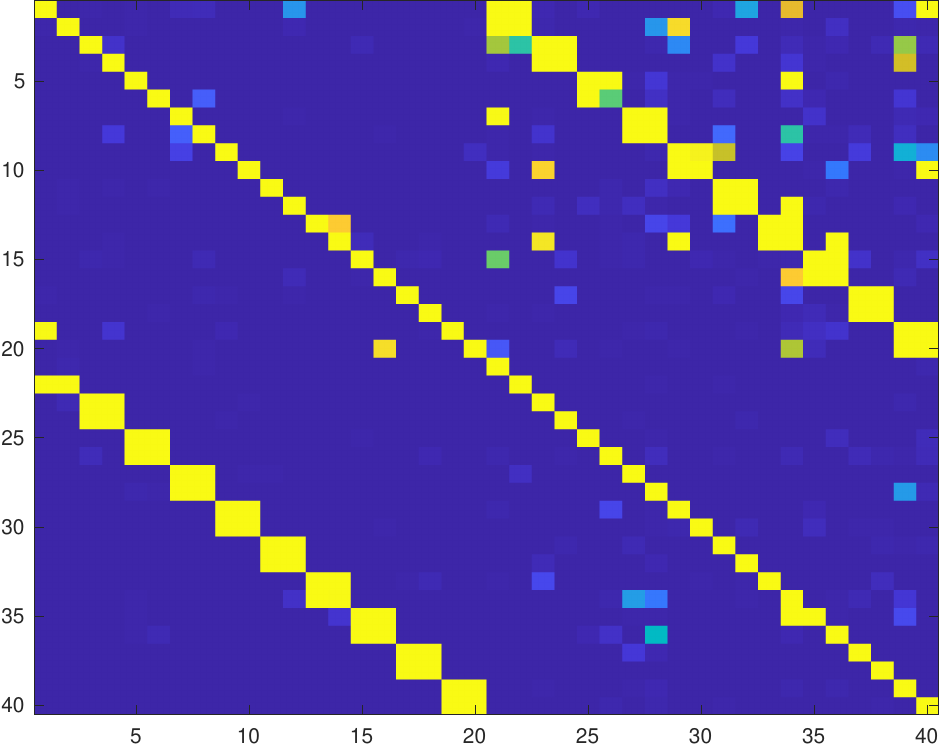}
         \end{subfigure}
          \begin{subfigure}[b]{0.32\linewidth}
                 \centering
                 \includegraphics[width=0.99\textwidth]{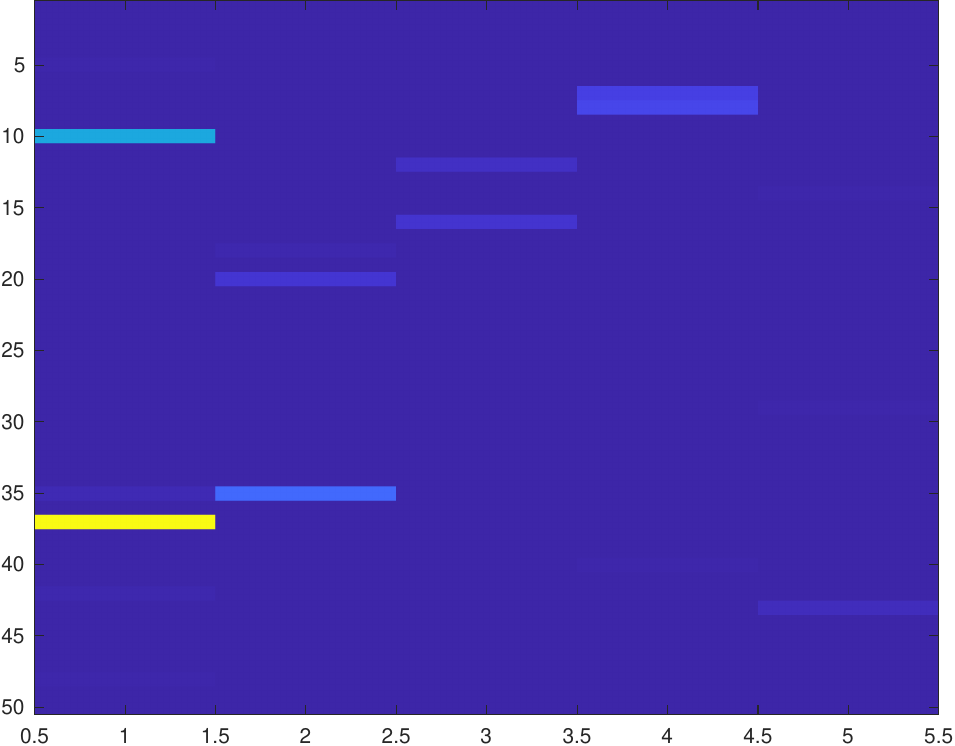}
         \end{subfigure}
       
\caption{Qualitative analysis of PFGCG ($V=1$, $\taumax=5$) on Lotka–Volterra. Left: Ground-truth GC graph, middle: Bernoulli posterior mean of the discovered GC graphs, right: Matrix of $\{r^{\tau}_k\}_{k=1, \tau}^{K, \taumax}$. Each rectangle in the figures indicates a value of a matrix and brighter colors indicates larger values.}
\label{fig-demo}
 \vspace{-0.2cm}
\end{wrapfigure}

In Figure~\ref{fig-demo}, we compare the Bernoulli posterior mean of PFGCG with the ground-truth graph on Lotka–Volterra. We can see that the posterior mean discovered by our method is well aligned with the ground-truth graph, where brighter rectangles indicate higher probability of a GC link between two variables. Finally, recall that $r^{\tau}_k$ in Eq.~(\ref{eq-base-dis}) models the weight of latent factor $k$ at lag $\tau$.
We plot $\{r^{\tau}_k\}_{k=1, \tau}^{K, \taumax}$ as a $K \times \taumax$ matrix. It can be seen that the matrix is quite sparse, where only a few entries have large values, i.e., only a few factors are active among $K=50$. This demonstrates the shrinkage mechanism of PFGCG on $K$.

\subsection{Qualitative Analysis on Climate Data}

\begin{figure}[t]
\centering
\captionsetup[subfigure]{justification=centering}
        \centering
         \begin{subfigure}[b]{0.69\linewidth}
                 \centering
                 \includegraphics[width=0.99\textwidth]{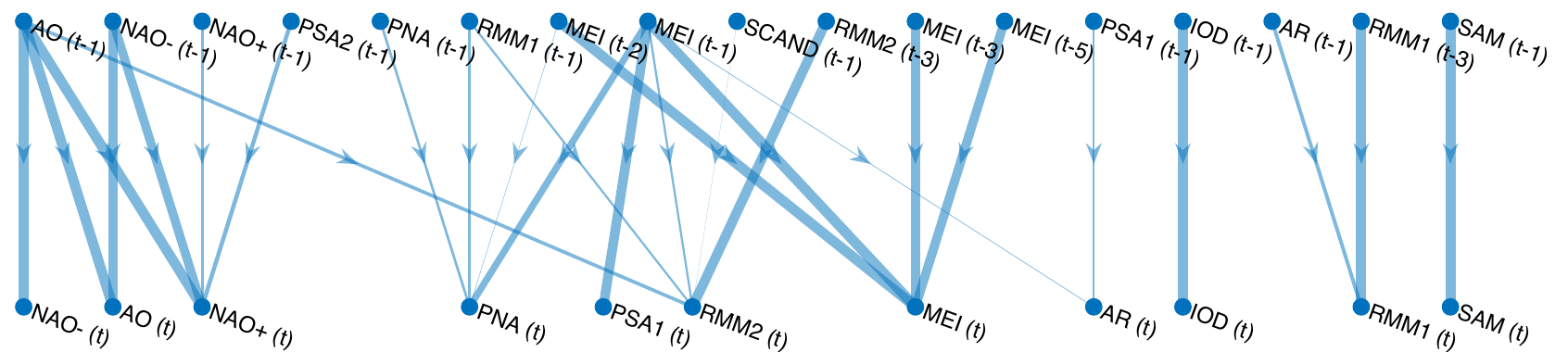}
         \end{subfigure}
         \begin{subfigure}[b]{0.29\linewidth}
                 \centering
                 \includegraphics[width=0.99\textwidth]{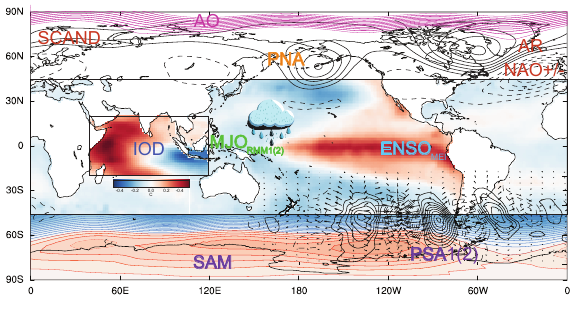}
         \end{subfigure}
\caption{Results on JR55. Left: Discovered casual links between indices on JR55 where the weights are from the Bernoulli posterior of the graph (links with weights less than 0.2 are not shown) and thicker links indicate stronger connections. Right: The geographical locations of the indices.}
\label{fig-climate-vis}
\end{figure}

We qualitatively analyse our method's performance on climate data obtained from the Japanese reanalysis of the atmosphere (JRA55)~\cite{kobayashi2015jra}. Following~\cite{harries2021dynamic}, we compute a set of 13 indices diagnosing the activity of the major atmospheric tropospheric and convective global climate modes at monthly resolution from 1960 to 2005, resulting in an MTS dataset where $N=13$ and $T=551$. More details of this dataset is provided in Section~\ref{sec-more-climate}.
We run our method with lag $\taumax=6$ and report the qualitative results in Figure~\ref{fig-climate-vis}. %
The resulting causal network reveals significant autocorrelation in the major convective modes that strongly couples to the ocean via surface temperature fluxes on interannual and intra-seasonal timescales i.e., the El Nino Southern Oscillation (ENSO) and the Madden-Julian Oscillation (MJO) respectively. The revealed posterior link weights closely correspond to the results discovered in~\cite{OKane2024Bayesian} on the same data (see their figure 8), a recent study of Bayesian structure learning on climate data that uses a different model and standard RJMCMC methods. Such an approach is several orders of magnitude more computationally expensive than the work presented here.

\subsection{Computational Complexity}
\label{sec-complexity}
\begin{figure*}[t]
\captionsetup[subfigure]{justification=centering}
        \centering
         \begin{subfigure}[b]{0.19\linewidth}
                 \centering
                 \includegraphics[width=0.99\textwidth]{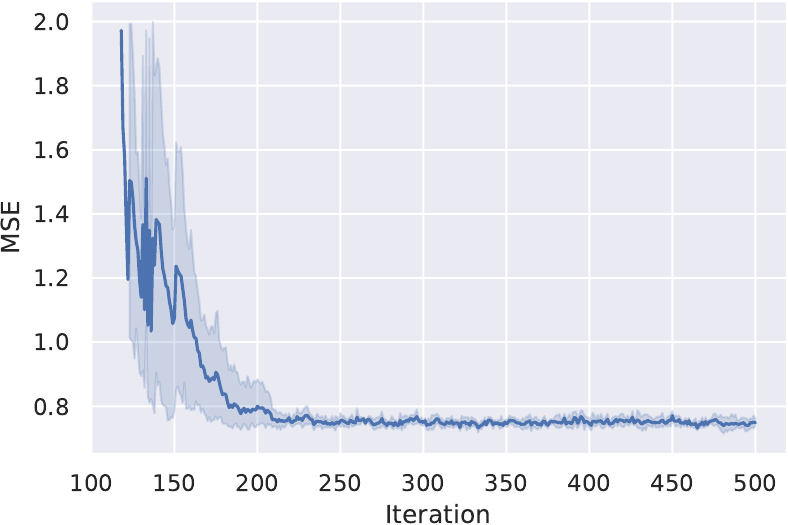}
                 \caption{L. 96, $T=100$}
         \end{subfigure}
         \begin{subfigure}[b]{0.19\linewidth}
                 \centering
                 \includegraphics[width=0.99\textwidth]{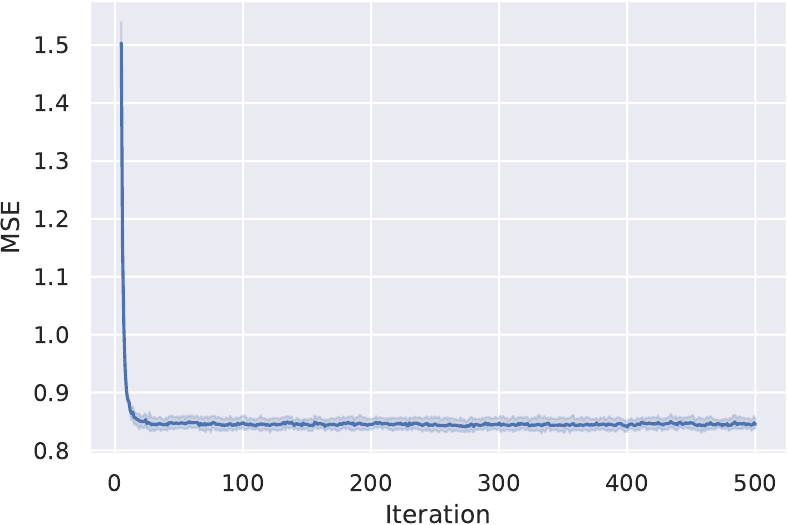}
                 \caption{L. 96, $T=500$}
         \end{subfigure}
          \begin{subfigure}[b]{0.19\linewidth}
                 \centering
                 \includegraphics[width=0.99\textwidth]{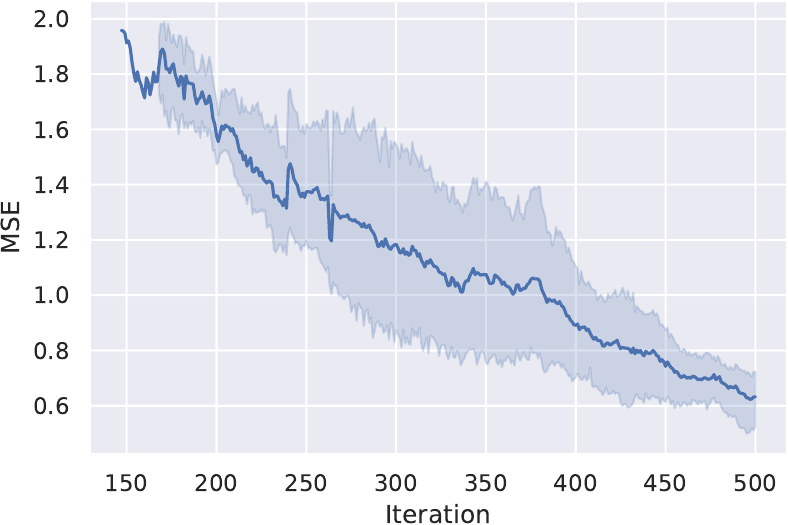}
                 \caption{L.–V., $T=200$}
         \end{subfigure}
       \begin{subfigure}[b]{0.19\linewidth}
                 \centering
                 \includegraphics[width=0.99\textwidth]{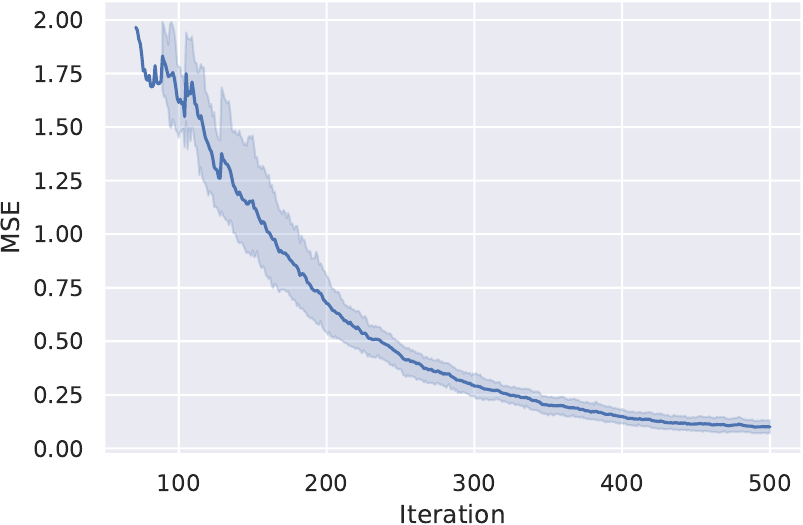}
                 \caption{L.–V., $T=500$}
         \end{subfigure}
          \begin{subfigure}[b]{0.19\linewidth}
                 \centering
                 \includegraphics[width=0.99\textwidth]{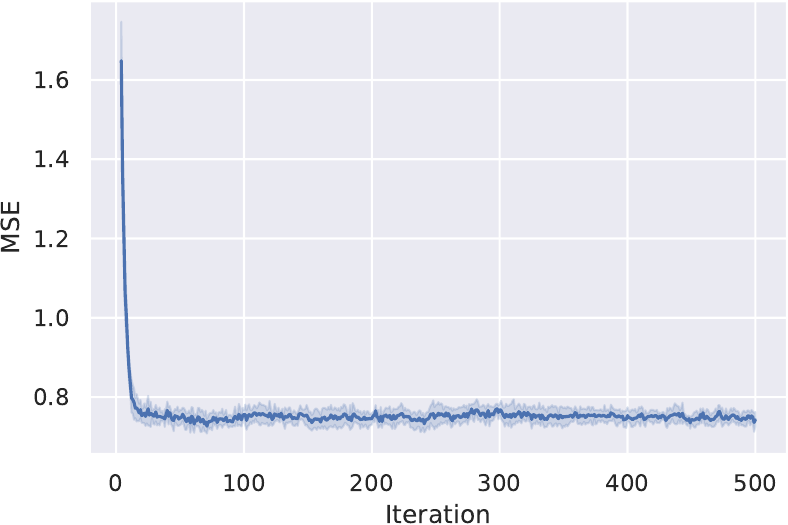}
                 \caption{FMRI}
         \end{subfigure}
       
\caption{MSE over iterations of PFGCG ($V=2$). For better visualisation, we show MSE in the iterations between $[a, b]$ where $a$ is the first iteration that MSE goes below 2.0 and $b=500$.}
\label{fig-mse}
 \vspace{-0.5cm}
\end{figure*}

\begin{wraptable}{r}{0.5\textwidth}
\centering
\caption{Running time (in seconds) per iteration on Lorenz 96 with $\taumax=3, V=2, K=50$.}
\resizebox{0.99\linewidth}{!}{
\begin{tabular}{@{}ccccc@{}} \toprule
        & $T=100$ & $T=500$ & $T=1,000$ & $T=5,000$ \\
$N=40$  & 0.23    & 0.23    & 0.27     & 0.72     \\
$N=128$ & 2.29    & 2.92    & 3.28     & 7.73     \\ \bottomrule
\end{tabular}
}
\label{tb-time}
\end{wraptable}

In addition to the complexity of our method analysed in Section~\ref{sec-inference-short}, we empirically study the computation complexity by examining two aspects: running time of one Gibbs sampling iteration and the number of iterations for convergence. 
For all the experiments here, our method ran on an Apple laptop with an M1 Pro processor. we We report the running time (seconds per Gibbs sampling iteration) of our method varying the number of variables and timestamps on Lorenz 96 in Table~\ref{tb-time}. It can be seen that the inference of our method is quite efficient. Figure~\ref{fig-mse} shows the MSE of PFGCG over the inference iterations. Although we set the maximum number of Gibbs sampling iterations to 10,000, in most cases our method converges around 200 iterations. 
As Lotka–Volterra has more complex dynamics, the model may need more iterations to converge (around 2,500 iterations). We can also see that in low-data regimes (e.g., $T=100$ on Lorenz 96 and $T=200$ on Lotka–Volterra), the samples of the model may have more variances in the beginning of the inference, which is as respected as there are less timestamps. In general, our method converges within half an hour on a laptop in most cases.

\section{Conclusion}
In this paper, we have presented a new Bayesian VAR model for Granger causal discovery on multivariate time-series data,
which imposes a Poisson factorised prior (PFGCG) modelling binary GC graphs separately from the VAR coefficients. PFGCG consists of a Poisson factorisation construction that generates count-valued  matrices, which are then used to generate binary GC graphs with the proposed link function (GBPL). Our model enjoys local conjugacy and Bayesian nonparametric properties, facilitating an efficient Bayesian inference algorithm with less hyperparameters. We have conducted extensive experiments comparing our method with Bayesian VARs and deep VARs, showing that ours has a significant performance advantage in low-data regimes. The limitation of the proposed method is that it follows the assumption of Granger causality and assumes causal relations take a factorisation form, which may not hold in other data in practice. Therefore, careful checks are needed when using the results of the method for decision making.

\bibliography{arxiv}

\begin{thebibliography}{78}
\providecommand{\natexlab}[1]{#1}
\providecommand{\url}[1]{\texttt{#1}}
\expandafter\ifx\csname urlstyle\endcsname\relax
  \providecommand{\doi}[1]{doi: #1}\else
  \providecommand{\doi}{doi: \begingroup \urlstyle{rm}\Url}\fi

\bibitem[Ahelegbey et~al.(2016)Ahelegbey, Billio, and Casarin]{ahelegbey2016sparse}
D.~F. Ahelegbey, M.~Billio, and R.~Casarin.
\newblock Sparse graphical vector autoregression: A bayesian approach.
\newblock \emph{Annals of Economics and Statistics/Annales d'{\'E}conomie et de Statistique}, \penalty0 (123/124):\penalty0 333--361, 2016.

\bibitem[Alvarez~Melis and Jaakkola(2018)]{alvarez2018towards}
D.~Alvarez~Melis and T.~Jaakkola.
\newblock Towards robust interpretability with self-explaining neural networks.
\newblock \emph{Advances in neural information processing systems}, 31, 2018.

\bibitem[Annadani et~al.(2023)Annadani, Pawlowski, Jennings, Bauer, Zhang, and Gong]{annadani2023bayesdag}
Y.~Annadani, N.~Pawlowski, J.~Jennings, S.~Bauer, C.~Zhang, and W.~Gong.
\newblock Bayesdag: Gradient-based posterior inference for causal discovery.
\newblock \emph{Advances in Neural Information Processing Systems}, 36, 2023.

\bibitem[Assaad et~al.(2022)Assaad, Devijver, and Gaussier]{assaad2022survey}
C.~K. Assaad, E.~Devijver, and E.~Gaussier.
\newblock Survey and evaluation of causal discovery methods for time series.
\newblock \emph{Journal of Artificial Intelligence Research}, 73:\penalty0 767--819, 2022.

\bibitem[Baca{\"e}r and Baca{\"e}r(2011)]{bacaer2011lotka}
N.~Baca{\"e}r and N.~Baca{\"e}r.
\newblock Lotka, volterra and the predator--prey system (1920--1926).
\newblock \emph{A short history of mathematical population dynamics}, pages 71--76, 2011.

\bibitem[Bengio et~al.(2023)Bengio, Lahlou, Deleu, Hu, Tiwari, and Bengio]{bengio2023gflownet}
Y.~Bengio, S.~Lahlou, T.~Deleu, E.~J. Hu, M.~Tiwari, and E.~Bengio.
\newblock Gflownet foundations.
\newblock \emph{Journal of Machine Learning Research}, 24\penalty0 (210):\penalty0 1--55, 2023.

\bibitem[Benjamini and Hochberg(1995)]{benjamini1995controlling}
Y.~Benjamini and Y.~Hochberg.
\newblock Controlling the false discovery rate: a practical and powerful approach to multiple testing.
\newblock \emph{Journal of the Royal statistical society: series B (Methodological)}, 57\penalty0 (1):\penalty0 289--300, 1995.

\bibitem[Billio et~al.(2019)Billio, Casarin, and Rossini]{billio2019bayesian}
M.~Billio, R.~Casarin, and L.~Rossini.
\newblock Bayesian nonparametric sparse var models.
\newblock \emph{Journal of Econometrics}, 212\penalty0 (1):\penalty0 97--115, 2019.

\bibitem[Breitung and Swanson(2002)]{breitung2002temporal}
J.~Breitung and N.~R. Swanson.
\newblock Temporal aggregation and spurious instantaneous causality in multiple time series models.
\newblock \emph{Journal of Time Series Analysis}, 23\penalty0 (6):\penalty0 651--665, 2002.

\bibitem[Bussmann et~al.(2021)Bussmann, Nys, and Latr{\'e}]{bussmann2021neural}
B.~Bussmann, J.~Nys, and S.~Latr{\'e}.
\newblock Neural additive vector autoregression models for causal discovery in time series.
\newblock In \emph{International Conference on Discovery Science}, pages 446--460. Springer, 2021.

\bibitem[Canny(2004)]{canny2004gap}
J.~Canny.
\newblock Gap: a factor model for discrete data.
\newblock In \emph{Proceedings of the 27th annual international ACM SIGIR conference on Research and development in information retrieval}, pages 122--129, 2004.

\bibitem[Casella and George(1992)]{casella1992explaining}
G.~Casella and E.~I. George.
\newblock Explaining the gibbs sampler.
\newblock \emph{The American Statistician}, 46\penalty0 (3):\penalty0 167--174, 1992.

\bibitem[Cheng et~al.(2023)Cheng, Yang, Xiao, Li, Suo, He, and Dai]{cheng2023cuts}
Y.~Cheng, R.~Yang, T.~Xiao, Z.~Li, J.~Suo, K.~He, and Q.~Dai.
\newblock Cuts: Neural causal discovery from irregular time-series data.
\newblock In \emph{The Eleventh International Conference on Learning Representations}, 2023.

\bibitem[Cheng et~al.(2024{\natexlab{a}})Cheng, Li, Xiao, Li, Suo, He, and Dai]{cheng2024cuts+}
Y.~Cheng, L.~Li, T.~Xiao, Z.~Li, J.~Suo, K.~He, and Q.~Dai.
\newblock Cuts+: High-dimensional causal discovery from irregular time-series.
\newblock In \emph{Proceedings of the AAAI Conference on Artificial Intelligence}, volume~38, pages 11525--11533, 2024{\natexlab{a}}.

\bibitem[Cheng et~al.(2024{\natexlab{b}})Cheng, Wang, Xiao, Zhong, Suo, and He]{cheng2024causaltime}
Y.~Cheng, Z.~Wang, T.~Xiao, Q.~Zhong, J.~Suo, and K.~He.
\newblock Causaltime: Realistically generated time-series for benchmarking of causal discovery.
\newblock In \emph{The Twelfth International Conference on Learning Representations}, 2024{\natexlab{b}}.

\bibitem[Dean and Kanazawa(1989)]{dean1989model}
T.~Dean and K.~Kanazawa.
\newblock A model for reasoning about persistence and causation.
\newblock \emph{Computational intelligence}, 5\penalty0 (2):\penalty0 142--150, 1989.

\bibitem[Deleu et~al.(2023)Deleu, Nishikawa-Toomey, Subramanian, Malkin, Charlin, and Bengio]{deleu2023joint}
T.~Deleu, M.~Nishikawa-Toomey, J.~Subramanian, N.~Malkin, L.~Charlin, and Y.~Bengio.
\newblock Joint bayesian inference of graphical structure and parameters with a single generative flow network.
\newblock \emph{Advances in Neural Information Processing Systems}, 36, 2023.

\bibitem[Demiralp and Hoover(2003)]{demiralp2003searching}
S.~Demiralp and K.~D. Hoover.
\newblock Searching for the causal structure of a vector autoregression.
\newblock \emph{Oxford Bulletin of Economics and statistics}, 65:\penalty0 745--767, 2003.

\bibitem[Fan et~al.(2023)Fan, Wang, Zhang, and Ouyang]{fan2023interpretable}
C.~Fan, Y.~Wang, Y.~Zhang, and W.~Ouyang.
\newblock Interpretable multi-scale neural network for granger causality discovery.
\newblock In \emph{ICASSP 2023-2023 IEEE International Conference on Acoustics, Speech and Signal Processing (ICASSP)}, pages 1--5. IEEE, 2023.

\bibitem[Ferguson(1973)]{ferguson1973bayesian}
T.~S. Ferguson.
\newblock A {B}ayesian analysis of some nonparametric problems.
\newblock \emph{The annals of statistics}, pages 209--230, 1973.

\bibitem[Fox et~al.(2011)Fox, Sudderth, Jordan, and Willsky]{fox2011bayesian}
E.~Fox, E.~B. Sudderth, M.~I. Jordan, and A.~S. Willsky.
\newblock Bayesian nonparametric inference of switching dynamic linear models.
\newblock \emph{IEEE Transactions on signal processing}, 59\penalty0 (4):\penalty0 1569--1585, 2011.

\bibitem[Geffner et~al.(2022)Geffner, Antoran, Foster, Gong, Ma, Kiciman, Sharma, Lamb, Kukla, Pawlowski, et~al.]{geffner2022deep}
T.~Geffner, J.~Antoran, A.~Foster, W.~Gong, C.~Ma, E.~Kiciman, A.~Sharma, A.~Lamb, M.~Kukla, N.~Pawlowski, et~al.
\newblock Deep end-to-end causal inference.
\newblock \emph{arXiv preprint arXiv:2202.02195}, 2022.

\bibitem[George et~al.(2008)George, Sun, and Ni]{george2008bayesian}
E.~I. George, D.~Sun, and S.~Ni.
\newblock Bayesian stochastic search for var model restrictions.
\newblock \emph{Journal of Econometrics}, 142\penalty0 (1):\penalty0 553--580, 2008.

\bibitem[Gershman and Blei(2012)]{gershman2012tutorial}
S.~J. Gershman and D.~M. Blei.
\newblock A tutorial on bayesian nonparametric models.
\newblock \emph{Journal of Mathematical Psychology}, 56\penalty0 (1):\penalty0 1--12, 2012.

\bibitem[Ghosh et~al.(2018)Ghosh, Khare, and Michailidis]{ghosh2018high}
S.~Ghosh, K.~Khare, and G.~Michailidis.
\newblock High-dimensional posterior consistency in bayesian vector autoregressive models.
\newblock \emph{Journal of the American Statistical Association}, 2018.

\bibitem[Giles(2016)]{giles2016algorithm}
M.~B. Giles.
\newblock Algorithm 955: approximation of the inverse poisson cumulative distribution function.
\newblock \emph{ACM transactions on mathematical software (TOMS)}, 42\penalty0 (1):\penalty0 1--22, 2016.

\bibitem[Gong et~al.(2023)Gong, Yao, Zhang, Li, Bi, Du, and Wang]{gong2023causal}
C.~Gong, D.~Yao, C.~Zhang, W.~Li, J.~Bi, L.~Du, and J.~Wang.
\newblock Causal discovery from temporal data.
\newblock In \emph{Proceedings of the 29th ACM SIGKDD Conference on Knowledge Discovery and Data Mining}, pages 5803--5804, 2023.

\bibitem[Gong et~al.(2022)Gong, Jennings, Zhang, and Pawlowski]{gong2022rhino}
W.~Gong, J.~Jennings, C.~Zhang, and N.~Pawlowski.
\newblock Rhino: Deep causal temporal relationship learning with history-dependent noise.
\newblock \emph{arXiv preprint arXiv:2210.14706}, 2022.

\bibitem[Gopalan et~al.(2014)Gopalan, Charlin, and Blei]{gopalan2014content}
P.~K. Gopalan, L.~Charlin, and D.~Blei.
\newblock Content-based recommendations with poisson factorization.
\newblock \emph{Advances in neural information processing systems}, 27, 2014.

\bibitem[Granger(1969)]{granger1969investigating}
C.~W. Granger.
\newblock Investigating causal relations by econometric models and cross-spectral methods.
\newblock \emph{Econometrica: journal of the Econometric Society}, pages 424--438, 1969.

\bibitem[Guo et~al.(2017)Guo, Pleiss, Sun, and Weinberger]{guo2017calibration}
C.~Guo, G.~Pleiss, Y.~Sun, and K.~Q. Weinberger.
\newblock On calibration of modern neural networks.
\newblock In \emph{International conference on machine learning}, pages 1321--1330. PMLR, 2017.

\bibitem[Harries and O'Kane(2021)]{harries2021dynamic}
D.~Harries and T.~J. O'Kane.
\newblock Dynamic bayesian networks for evaluation of granger causal relationships in climate reanalyses.
\newblock \emph{Journal of Advances in Modeling Earth Systems}, 13\penalty0 (5):\penalty0 e2020MS002442, 2021.

\bibitem[Huang et~al.(2020)Huang, Zhang, Zhang, Ramsey, Sanchez-Romero, Glymour, and Sch{\"o}lkopf]{huang2020causal}
B.~Huang, K.~Zhang, J.~Zhang, J.~D. Ramsey, R.~Sanchez-Romero, C.~Glymour, and B.~Sch{\"o}lkopf.
\newblock Causal discovery from heterogeneous/nonstationary data.
\newblock \emph{J. Mach. Learn. Res.}, 21\penalty0 (89):\penalty0 1--53, 2020.

\bibitem[Hyv{\"a}rinen et~al.(2010)Hyv{\"a}rinen, Zhang, Shimizu, and Hoyer]{hyvarinen2010estimation}
A.~Hyv{\"a}rinen, K.~Zhang, S.~Shimizu, and P.~O. Hoyer.
\newblock Estimation of a structural vector autoregression model using non-gaussianity.
\newblock \emph{Journal of Machine Learning Research}, 11\penalty0 (5), 2010.

\bibitem[Karimi and Paul(2010)]{karimi2010extensive}
A.~Karimi and M.~R. Paul.
\newblock Extensive chaos in the lorenz-96 model.
\newblock \emph{Chaos: An interdisciplinary journal of nonlinear science}, 20\penalty0 (4), 2010.

\bibitem[Khanna and Tan(2020)]{khanna2019economy}
S.~Khanna and V.~Y. Tan.
\newblock Economy statistical recurrent units for inferring nonlinear granger causality.
\newblock In \emph{International Conference on Learning Representations}, 2020.

\bibitem[Kobayashi et~al.(2015)Kobayashi, Ota, Harada, Ebita, Moriya, Onoda, Onogi, Kamahori, Kobayashi, Endo, et~al.]{kobayashi2015jra}
S.~Kobayashi, Y.~Ota, Y.~Harada, A.~Ebita, M.~Moriya, H.~Onoda, K.~Onogi, H.~Kamahori, C.~Kobayashi, H.~Endo, et~al.
\newblock The jra-55 reanalysis: General specifications and basic characteristics.
\newblock \emph{Journal of the Meteorological Society of Japan. Ser. II}, 93\penalty0 (1):\penalty0 5--48, 2015.

\bibitem[Kumar et~al.(2019)Kumar, Liang, and Ma]{kumar2019verified}
A.~Kumar, P.~S. Liang, and T.~Ma.
\newblock Verified uncertainty calibration.
\newblock \emph{Advances in Neural Information Processing Systems}, 32, 2019.

\bibitem[Litterman(1986)]{litterman1986forecasting}
R.~B. Litterman.
\newblock Forecasting with bayesian vector autoregressions—five years of experience.
\newblock \emph{Journal of Business \& Economic Statistics}, 4\penalty0 (1):\penalty0 25--38, 1986.

\bibitem[Liu et~al.(2020)Liu, Lin, Padhy, Tran, Bedrax~Weiss, and Lakshminarayanan]{liu2020simple}
J.~Liu, Z.~Lin, S.~Padhy, D.~Tran, T.~Bedrax~Weiss, and B.~Lakshminarayanan.
\newblock Simple and principled uncertainty estimation with deterministic deep learning via distance awareness.
\newblock \emph{Advances in neural information processing systems}, 33:\penalty0 7498--7512, 2020.

\bibitem[Lorenz(1996)]{lorenz1996predictability}
E.~N. Lorenz.
\newblock Predictability: A problem partly solved.
\newblock In \emph{Proc. Seminar on predictability}, volume~1. Reading, 1996.

\bibitem[L{\"o}we et~al.(2022)L{\"o}we, Madras, Zemel, and Welling]{lowe2022amortized}
S.~L{\"o}we, D.~Madras, R.~Zemel, and M.~Welling.
\newblock Amortized causal discovery: Learning to infer causal graphs from time-series data.
\newblock In \emph{Conference on Causal Learning and Reasoning}, pages 509--525. PMLR, 2022.

\bibitem[Lozano et~al.(2009)Lozano, Abe, Liu, and Rosset]{lozano2009grouped}
A.~C. Lozano, N.~Abe, Y.~Liu, and S.~Rosset.
\newblock Grouped graphical granger modeling for gene expression regulatory networks discovery.
\newblock \emph{Bioinformatics}, 25\penalty0 (12):\penalty0 i110--i118, 2009.

\bibitem[L{\"u}tkepohl(2005)]{lutkepohl2005new}
H.~L{\"u}tkepohl.
\newblock \emph{New introduction to multiple time series analysis}.
\newblock Springer Science \& Business Media, 2005.

\bibitem[Marcinkevi{\v{c}}s and Vogt(2021)]{marcinkevivcs2020interpretable}
R.~Marcinkevi{\v{c}}s and J.~E. Vogt.
\newblock Interpretable models for granger causality using self-explaining neural networks.
\newblock In \emph{International Conference on Learning Representations}, 2021.

\bibitem[Mihajlovic and Petkovic(2001)]{mihajlovic2001dynamic}
V.~Mihajlovic and M.~Petkovic.
\newblock Dynamic bayesian networks: A state of the art.
\newblock \emph{University of Twente Document Repository}, 2001.

\bibitem[Miranda-Agrippino and Ricco(2019)]{miranda2019bayesian}
S.~Miranda-Agrippino and G.~Ricco.
\newblock Bayesian vector autoregressions: Estimation.
\newblock In \emph{Oxford Research Encyclopedia of Economics and Finance}. 2019.

\bibitem[Montalto et~al.(2015)Montalto, Stramaglia, Faes, Tessitore, Prevete, and Marinazzo]{montalto2015neural}
A.~Montalto, S.~Stramaglia, L.~Faes, G.~Tessitore, R.~Prevete, and D.~Marinazzo.
\newblock Neural networks with non-uniform embedding and explicit validation phase to assess granger causality.
\newblock \emph{Neural networks}, 71:\penalty0 159--171, 2015.

\bibitem[Murphy(2002)]{murphy2002dynamic}
K.~P. Murphy.
\newblock \emph{Dynamic bayesian networks: representation, inference and learning}.
\newblock University of California, Berkeley, 2002.

\bibitem[Nakajima and West(2013)]{nakajima2013bayesian}
J.~Nakajima and M.~West.
\newblock Bayesian analysis of latent threshold dynamic models.
\newblock \emph{Journal of Business \& Economic Statistics}, 31\penalty0 (2):\penalty0 151--164, 2013.

\bibitem[Nauta et~al.(2019)Nauta, Bucur, and Seifert]{nauta2019causal}
M.~Nauta, D.~Bucur, and C.~Seifert.
\newblock Causal discovery with attention-based convolutional neural networks.
\newblock \emph{Machine Learning and Knowledge Extraction}, 1\penalty0 (1):\penalty0 19, 2019.

\bibitem[Nicholson et~al.(2017)Nicholson, Matteson, and Bien]{nicholson2017varx}
W.~B. Nicholson, D.~S. Matteson, and J.~Bien.
\newblock Varx-l: Structured regularization for large vector autoregressions with exogenous variables.
\newblock \emph{International Journal of Forecasting}, 33\penalty0 (3):\penalty0 627--651, 2017.

\bibitem[O'Kane et~al.(2024)O'Kane, Harries, and Collier]{OKane2024Bayesian}
T.~J. O'Kane, D.~Harries, and M.~A. Collier.
\newblock Bayesian structure learning for climate model evaluation.
\newblock \emph{Journal of Advances in Modeling Earth Systems}, 16\penalty0 (5):\penalty0 e2023MS004034, 2024.

\bibitem[Oliva et~al.(2017)Oliva, P{\'o}czos, and Schneider]{oliva2017statistical}
J.~B. Oliva, B.~P{\'o}czos, and J.~Schneider.
\newblock The statistical recurrent unit.
\newblock In \emph{International Conference on Machine Learning}, pages 2671--2680. PMLR, 2017.

\bibitem[Orbanz and Teh(2010)]{orbanz2010bayesian}
P.~Orbanz and Y.~W. Teh.
\newblock Bayesian nonparametric models.
\newblock \emph{Encyclopedia of machine learning}, 1:\penalty0 81--89, 2010.

\bibitem[Pamfil et~al.(2020)Pamfil, Sriwattanaworachai, Desai, Pilgerstorfer, Georgatzis, Beaumont, and Aragam]{pamfil2020dynotears}
R.~Pamfil, N.~Sriwattanaworachai, S.~Desai, P.~Pilgerstorfer, K.~Georgatzis, P.~Beaumont, and B.~Aragam.
\newblock Dynotears: Structure learning from time-series data.
\newblock In \emph{International Conference on Artificial Intelligence and Statistics}, pages 1595--1605. PMLR, 2020.

\bibitem[Peters et~al.(2013)Peters, Janzing, and Sch{\"o}lkopf]{peters2013causal}
J.~Peters, D.~Janzing, and B.~Sch{\"o}lkopf.
\newblock Causal inference on time series using restricted structural equation models.
\newblock \emph{Advances in Neural Information Processing Systems}, 26, 2013.

\bibitem[Runge(2018)]{runge2018causal}
J.~Runge.
\newblock Causal network reconstruction from time series: From theoretical assumptions to practical estimation.
\newblock \emph{Chaos: An Interdisciplinary Journal of Nonlinear Science}, 28\penalty0 (7):\penalty0 075310, 2018.

\bibitem[Runge(2020)]{runge2020discovering}
J.~Runge.
\newblock Discovering contemporaneous and lagged causal relations in autocorrelated nonlinear time series datasets.
\newblock In \emph{Conference on Uncertainty in Artificial Intelligence}, pages 1388--1397. PMLR, 2020.

\bibitem[Runge et~al.(2019)Runge, Nowack, Kretschmer, Flaxman, and Sejdinovic]{runge2019detecting}
J.~Runge, P.~Nowack, M.~Kretschmer, S.~Flaxman, and D.~Sejdinovic.
\newblock Detecting and quantifying causal associations in large nonlinear time series datasets.
\newblock \emph{Science advances}, 5\penalty0 (11):\penalty0 eaau4996, 2019.

\bibitem[Seabold and Statsmodels(2010)]{seaboldeconometric}
S.~Seabold and P.~Statsmodels.
\newblock Econometric and statistical modeling with python.
\newblock In \emph{Proceedings of the 9th Python in science conference}, pages 57--61, 2010.

\bibitem[Shiguihara et~al.(2021)Shiguihara, Lopes, and Mauricio]{shiguihara2021dynamic}
P.~Shiguihara, A.~D.~A. Lopes, and D.~Mauricio.
\newblock Dynamic bayesian network modeling, learning, and inference: a survey.
\newblock \emph{IEEE Access}, 9:\penalty0 117639--117648, 2021.

\bibitem[Shojaie and Fox(2022)]{shojaie2022granger}
A.~Shojaie and E.~B. Fox.
\newblock Granger causality: A review and recent advances.
\newblock \emph{Annual Review of Statistics and Its Application}, 9:\penalty0 289--319, 2022.

\bibitem[Smith et~al.(2011)Smith, Miller, Salimi-Khorshidi, Webster, Beckmann, Nichols, Ramsey, and Woolrich]{smith2011network}
S.~M. Smith, K.~L. Miller, G.~Salimi-Khorshidi, M.~Webster, C.~F. Beckmann, T.~E. Nichols, J.~D. Ramsey, and M.~W. Woolrich.
\newblock Network modelling methods for fmri.
\newblock \emph{Neuroimage}, 54\penalty0 (2):\penalty0 875--891, 2011.

\bibitem[Spirtes et~al.(2000)Spirtes, Glymour, Scheines, and Heckerman]{spirtes2000causation}
P.~Spirtes, C.~N. Glymour, R.~Scheines, and D.~Heckerman.
\newblock \emph{Causation, prediction, and search}.
\newblock MIT press, 2000.

\bibitem[Swanson and Granger(1997)]{swanson1997impulse}
N.~R. Swanson and C.~W. Granger.
\newblock Impulse response functions based on a causal approach to residual orthogonalization in vector autoregressions.
\newblock \emph{Journal of the American Statistical Association}, 92\penalty0 (437):\penalty0 357--367, 1997.

\bibitem[Tank et~al.(2018)Tank, Covert, Foti, Shojaie, and Fox]{tank2018neural}
A.~Tank, I.~Covert, N.~Foti, A.~Shojaie, and E.~Fox.
\newblock Neural granger causality for nonlinear time series.
\newblock \emph{stat}, 1050:\penalty0 16, 2018.

\bibitem[Tong et~al.(2022)Tong, Atanackovic, Hartford, and Bengio]{tong2022bayesian}
A.~Tong, L.~Atanackovic, J.~Hartford, and Y.~Bengio.
\newblock Bayesian dynamic causal discovery.
\newblock In \emph{A causal view on dynamical systems, NeurIPS 2022 workshop}, 2022.

\bibitem[Wang et~al.(2018)Wang, Lin, Qi, Lian, Feng, Wu, and Pan]{wang2018estimating}
Y.~Wang, K.~Lin, Y.~Qi, Q.~Lian, S.~Feng, Z.~Wu, and G.~Pan.
\newblock Estimating brain connectivity with varying-length time lags using a recurrent neural network.
\newblock \emph{IEEE Transactions on Biomedical Engineering}, 65\penalty0 (9):\penalty0 1953--1963, 2018.

\bibitem[Wolpert et~al.(2011)Wolpert, Clyde, Tu, et~al.]{wolpert2011stochastic}
R.~L. Wolpert, M.~A. Clyde, C.~Tu, et~al.
\newblock Stochastic expansions using continuous dictionaries: {L}{\'e}vy adaptive regression kernels.
\newblock \emph{The Annals of Statistics}, 39\penalty0 (4):\penalty0 1916--1962, 2011.

\bibitem[Wo{\'z}niak(2016)]{wozniak2016bayesian}
T.~Wo{\'z}niak.
\newblock Bayesian vector autoregressions.
\newblock \emph{Australian Economic Review}, 49\penalty0 (3):\penalty0 365--380, 2016.

\bibitem[Wu et~al.(2020)Wu, Breuel, Skuhersky, and Kautz]{wu2020discovering}
T.~Wu, T.~Breuel, M.~Skuhersky, and J.~Kautz.
\newblock Discovering nonlinear relations with minimum predictive information regularization.
\newblock \emph{arXiv preprint arXiv:2001.01885}, 2020.

\bibitem[Yang and Leskovec(2012)]{yang2012community}
J.~Yang and J.~Leskovec.
\newblock Community-affiliation graph model for overlapping network community detection.
\newblock In \emph{2012 IEEE 12th international conference on data mining}, pages 1170--1175. IEEE, 2012.

\bibitem[Yang and Leskovec(2014)]{yang2014structure}
J.~Yang and J.~Leskovec.
\newblock Structure and overlaps of ground-truth communities in networks.
\newblock \emph{ACM Transactions on Intelligent Systems and Technology (TIST)}, 5\penalty0 (2):\penalty0 1--35, 2014.

\bibitem[Yuan and Lin(2006)]{yuan2006model}
M.~Yuan and Y.~Lin.
\newblock Model selection and estimation in regression with grouped variables.
\newblock \emph{Journal of the Royal Statistical Society Series B: Statistical Methodology}, 68\penalty0 (1):\penalty0 49--67, 2006.

\bibitem[Zhou(2015)]{zhou2015infinite}
M.~Zhou.
\newblock Infinite edge partition models for overlapping community detection and link prediction.
\newblock In \emph{AISTATS}, pages 1135--1143, 2015.

\bibitem[Zhou and Carin(2013)]{zhou2013negative}
M.~Zhou and L.~Carin.
\newblock Negative binomial process count and mixture modeling.
\newblock \emph{IEEE Transactions on Pattern Analysis and Machine Intelligence}, 37\penalty0 (2):\penalty0 307--320, 2013.

\bibitem[Zhou et~al.(2012)Zhou, Hannah, Dunson, and Carin]{zhou2012beta}
M.~Zhou, L.~Hannah, D.~Dunson, and L.~Carin.
\newblock Beta-negative binomial process and poisson factor analysis.
\newblock In \emph{Artificial Intelligence and Statistics}, pages 1462--1471. PMLR, 2012.

\end{thebibliography}
\bibliographystyle{abbrvnat}

\appendix
\section{Appendix}

\subsection{Inference via Gibbs Sampling}
\label{sec-inference}
\paragraph{Sampling $\mA^\tau$}
With the conjugacy of normal distributions, one can sample the entries of $\mA^\tau$ one by one by:
\begin{align}
 A^\tau_{i,j} \sim 
\begin{dcases}
\pnormal{0}{(\psi^\tau_{i,j})^{-1}}, & \text{if } G^{\tau}_{i,j}=0\\
\pnormal{\mu^\tau_{i,j}}{\sigma^\tau_{i,j}},              & \text{otherwise}
\end{dcases}
\end{align}
where:
\begin{align}
\sigma^\tau_{i,j} = \left(\lambda_iG^\tau_{i,j}\sum_{t=1}^T x^2_{j, t-\tau} + \psi^\tau_{i,j}\right)^{-1},\\
\mu^\tau_{i,j} = \sigma^\tau_{i,j} G^\tau_{i,j}\lambda_i \left(\sum_{t=1}^T x^{\neg \tau, \neg j}_{i,t} x_{j, t-\tau}\right),\\
x^{\neg \tau, \neg j}_{i,t} = x_{i,t} - \sum_{j' \neq j}^N A^\tau_{i,j'} G^{\tau}_{i,j'} x_{j', t-\tau} - \sum_{\tau' \neq \tau}^{\taumax} \sum_{j'=1}^{N} A^{\tau'}_{i,j'} G^{\tau'}_{i,j'} x_{j', t-\tau'}
\end{align}

\paragraph{Sampling $\psi^\tau_{i,j}$ and $\lambda_i$}
With the conjugacy between normal and gamma distributions, one can sample $\psi^\tau_{i,j}$ and $\lambda_i$ from their conditional gamma posteriors:
\begin{align}
\psi^\tau_{i,j} \sim \pgamma{1.5}{1/(A^\tau_{i,j}/2 + 1)},\\
\lambda_i \sim \pgamma{1+T/2}{\left(1 + \sum_{t=1}^T (x_{i,t} - \sum_{\tau=1}^{\taumax} \sum_{j=1}^N A^\tau_{i,j} G^\tau_{i,j} x_{j, t-\tau})\right)^{-1}}.
\end{align}

\paragraph{Sampling $\mM^\tau$} With GBPL, we can sample:
\begin{align}
M^\tau_{i,j} \sim 
\begin{dcases}
\pcat{V}{\left[\dots, \frac{\frac{e^{-q^\tau_{i,j}} (q^\tau_{i,j})^v}{v!}}{\sum_{v'=0}^{V} \frac{e^{-q^\tau_{i,j}} (q^\tau_{i,j})^{v'}}{v'!}},\dots\right]}, & \text{if } G^{\tau}_{i,j}=0\\
\ptpoisson{V}{q^\tau_{i,j}},              & \text{otherwise}
\end{dcases}
\label{eq-sample-m}
\end{align}
where $q^\tau_{i,j} = \sum_{k=1}^K \theta^\tau_{i,k} r^\tau_k \phi^\tau_{j,k}$.

\paragraph{Sampling $M^\tau_{i,j,k}$} With the relationships between Poisson and multinomial distributions, we can sample:
\begin{align}
    [\dots, m^\tau_{i,j,k},\dots] \sim \pmulti{K}{m^\tau_{i,j}; \left[\dots, \frac{q^\tau_{i,j,k}}{\sum_{k'=1}^K q^\tau_{i,j,k'}},\dots\right]},
\end{align}
where $q^\tau_{i,j,k} = \theta^\tau_{i,k} r^\tau_k \phi^\tau_{j,k}$.

\paragraph{Sampling $\theta^\tau_{i,k}$, $\phi^\tau_{j,k}$, and $r^\tau_k$}
\begin{align}
    \theta^\tau_{i,k} \sim \pgamma{a^\tau_i + \sum_{j=1}^N M^\tau_{i,j, k}}{\frac{1}{d^\tau_k + r^\tau_k  \sum_{j=1}^N \phi^\tau_{j, k}}}, \\
    \phi^\tau_{j,k} \sim \pgamma{b^\tau_j +  \sum_{i=1}^N M^\tau_{i,j, k}}{\frac{1}{e^\tau_k + r^\tau_k  \sum_{i=1}^N \theta^\tau_{i, k}}},\\
    r^\tau_{k} \sim \pgamma{1/K +  \sum_{i=1,j=1}^N M^\tau_{i,j, k}}{\frac{1}{c^\tau + \sum_{i=1,j=1}^N \theta^\tau_{i, k} \phi^\tau_{j, k}}}.
\end{align}

\paragraph{Sampling $a^\tau_i$ and $b^\tau_j$} By introducing auxiliary variables from the Chinese Restaurant Table (CRT) distribution~\cite{zhou2012beta,zhou2013negative}, we can sample:
\begin{align}
    l^\tau_{i,k} \sim \pcrt{\sum_{j=1}^N M^\tau_{i,j, k}}{a^\tau_i},\\
    a^\tau_i \sim \pgamma{1 + \sum_{k=1}^K l^\tau_{i, k}}{\frac{1}{1 + \sum_{k=1}^K \log (1 + r^\tau_k \sum_{j=1}^N \phi^\tau_{j, k}  /d^\tau_k)}},\\
    o^\tau_{j,k} \sim \pcrt{\sum_{i=1}^N M^\tau_{i,j, k}}{b^\tau_j},\\
    b^\tau_j \sim \pgamma{1 + \sum_{k=1}^K o^\tau_{j, k}}{\frac{1}{1 + \sum_{k=1}^K \log (1 + r^\tau_k \sum_{i=1}^N \theta^\tau_{i, k}  /e^\tau_k)}}.
\end{align}

\paragraph{Sampling $d^\tau_k$, $e^\tau_k$ and $c^\tau$}
\begin{align}
d^\tau_k \sim \pgamma{\sum_{i=1}^N a^\tau_i+1}{\frac{1}{\sum_{i=1}^N\theta^\tau_{i, k} + 1}},\\
e^\tau_k \sim \pgamma{\sum_{j=1}^N b^\tau_j+1}{\frac{1}{\sum_{i=1}^N \phi^\tau_{j, k} + 1}},\\
c^\tau \sim \pgamma{2}{\frac{1}{\sum_{k=1}^K r^\tau_k + 1}}.
\end{align}

\begin{figure}[t]
  \centering
  \begin{minipage}{0.9\textwidth}
    \centering
\begin{algorithm2e}[H]
\SetKwInOut{Input}{input}\SetKwInOut{Output}{output}
\Input{MTS data $\mX$, number of lags $\taumax$, hyperparameter $V$}
\Output{Posterior samples of $\{\mA^\tau\}^{\taumax}_\tau$ and $\{\mG^\tau\}^{\taumax}_\tau$}
Initialise all the variables\;

\While{Not converged}
{
        \For{$i=1\dots N$}
        {
            Sample $\lambda_i$;
        }
        \For{$\tau=1\dots \taumax$}
        {
            \For{$i=1\dots N$, $j=1\dots N$}
            {
                Sample $M^\tau_{i,j}$ and $M^\tau_{i,j,k}$;
            }
            Sample $c^\tau$;
            
            \For{$i=1\dots N$}
            {
                Sample $a^\tau_i$ and $b^\tau_i$;
            }
            \For{$k=1\dots K$}
            {
                Sample $d^\tau_k$, $e^\tau_k$, $r^\tau_k$;
            }
            \For{$i=1\dots N$}
            {
                \For{$k=1\dots K$}
                {
                    Sample $\theta^\tau_{i,k}$ and $\phi^\tau_{i,k}$;
                }
            }
             
            \For{$i=1\dots N$}
            {
                \For{$j=1\dots N$}
                {
                    Sample $A^\tau_{i,j}$, $\psi^\tau_{i,j}$, $G^\tau_{i,j}$;
                }
            }

        }
}
\caption{Inference Algorithm for PFGCG.}
\label{alg}
\end{algorithm2e}
\end{minipage}
\end{figure}

\begin{table}[]
\centering
\caption{Hyperparameter settings.}
\label{tb-setting}
\resizebox{0.9\linewidth}{!}{
\begin{tabular}{|c|c|c|c|c|c|c|c|}
\hline
Model     & $\taumax$ & \# hidden layers & \# hidden units & \# training epochs & Learning rate & Mini-batch size & Parameter space                                                                                                 \\ \hline
VAR (FBH) & \{1,3,5\} & NA               & NA              & NA                  & NA            & NA              & NA                                                                                                              \\ \hline
BVAR(d)   & \{1,3,5\} & NA               & NA              & NA                  & NA            & NA              & NA                                                                                                              \\ \hline
BVAR(c)   & \{1,3,5\} & NA               & NA              & NA                  & NA            & NA              & NA                                                                                                              \\ \hline
SRU       & NA        & 1                & 10              & 2000                & 1.0e-3        & 50              & \begin{tabular}[c]{@{}c@{}}$\mu_1 = [0.01, 0.05]$\\ $\mu_2 = [0.01, 0.05]$\\ $\mu_3 = [0.01, 1.0]$\end{tabular} \\ \hline
eSRU      & NA        & 2                & 10              & 2000                & 1.0e-3        & 50              & \begin{tabular}[c]{@{}c@{}}$\mu_1 = [0.01, 0.05]$\\ $\mu_2 = [0.01, 0.05]$\\ $\mu_3 = [0.01, 1.0]$\end{tabular} \\ \hline
GVAR      & \{1,3,5\} & 2                & 50              & 1,000               & 1.0e-4        & 64              & \begin{tabular}[c]{@{}c@{}}$\lambda = [0.0, 3.0]$\\ $\gamma= [0.0, 0.1]$\end{tabular}                           \\ \hline
PFGCG     & \{1,3,5\} & NA               & NA              & 10,000              & NA            & NA              & $V = \{1,3,5\}$                                                                                                 \\ \hline
\end{tabular}
}
\end{table}

\begin{figure*}[t]

\captionsetup[subfigure]{justification=centering}

        \centering
         \begin{subfigure}[b]{0.14\linewidth}
                 \centering
                 \includegraphics[width=0.99\textwidth]{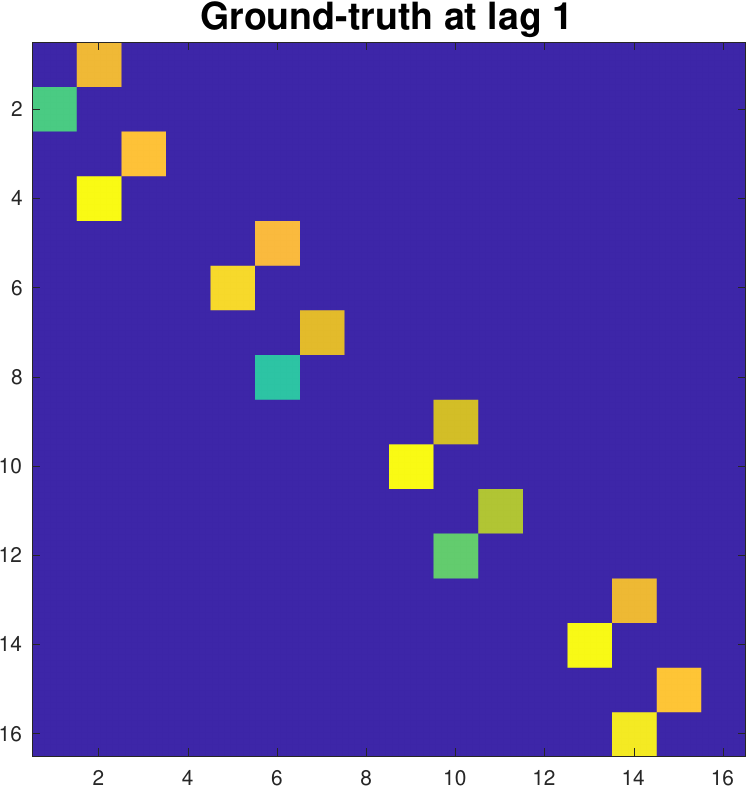}
         \end{subfigure}
         \begin{subfigure}[b]{0.14\linewidth}
                 \centering
                 \includegraphics[width=0.99\textwidth]{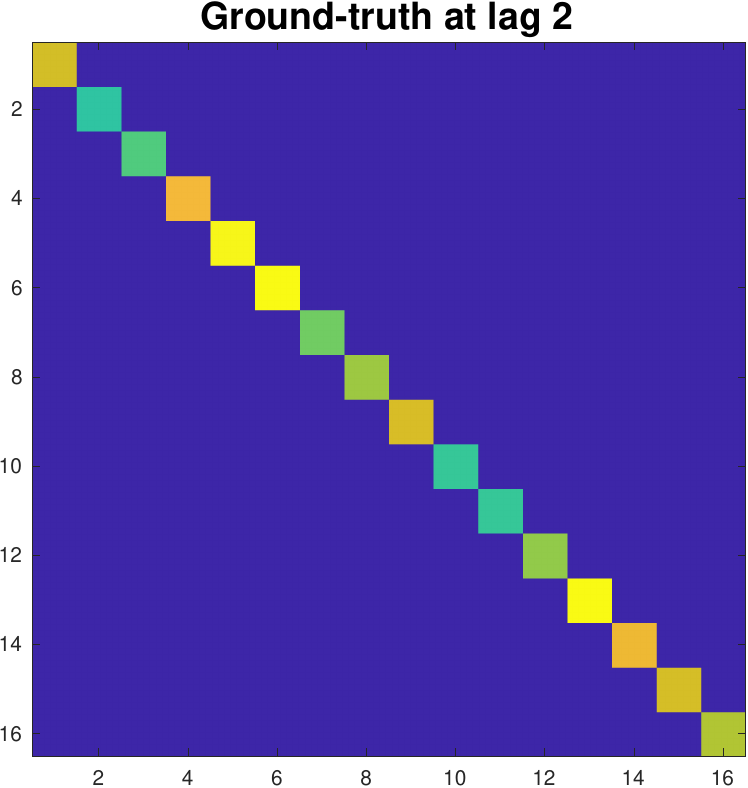}
         \end{subfigure} 
        \begin{subfigure}[b]{0.14\linewidth}
                 \centering
                 \includegraphics[width=0.99\textwidth]{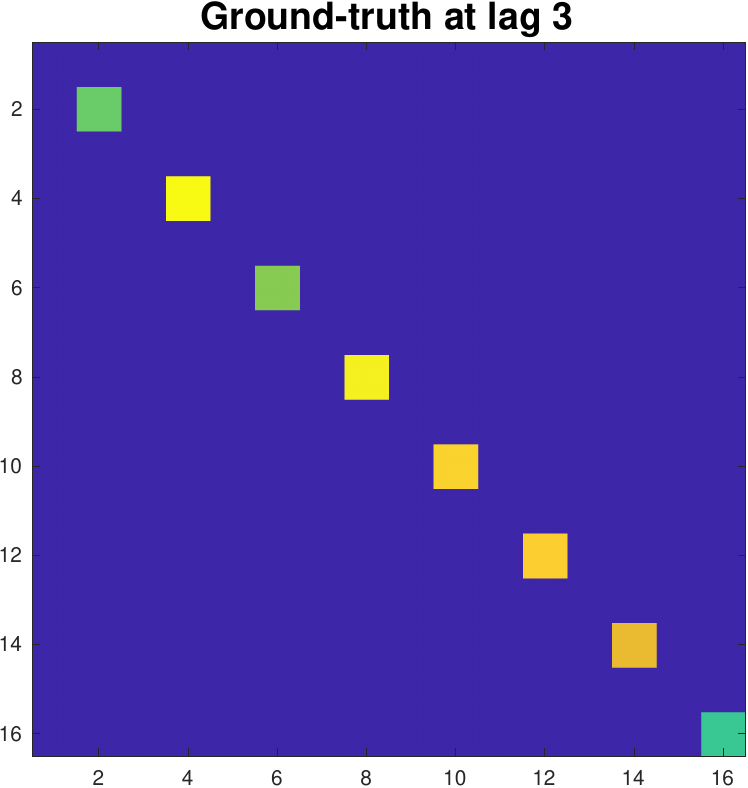}
         \end{subfigure}
    \begin{subfigure}[b]{0.14\linewidth}
                 \centering
                 \includegraphics[width=0.99\textwidth]{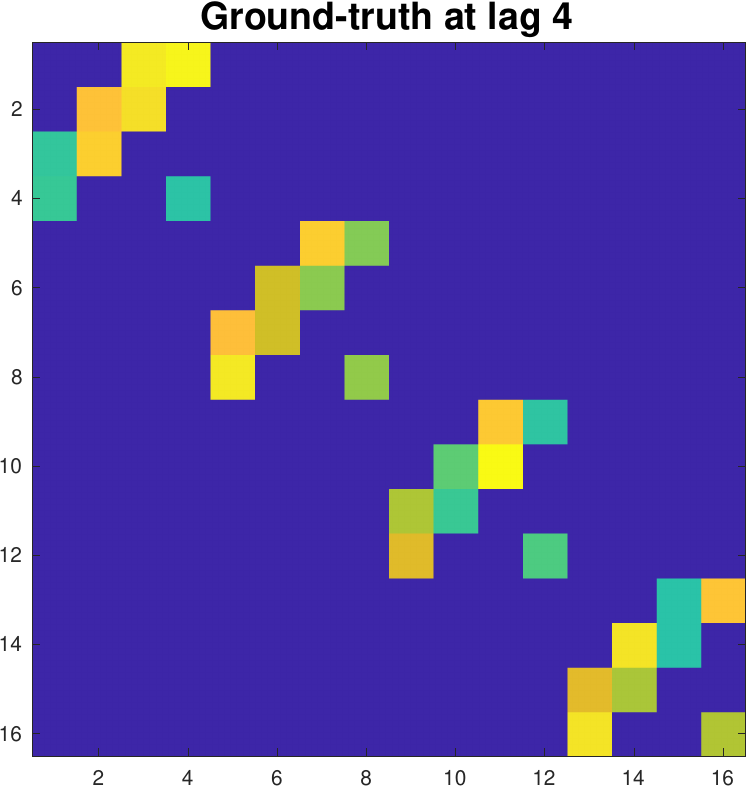}
         \end{subfigure}
         \begin{subfigure}[b]{0.14\linewidth}
                 \centering
                 \includegraphics[width=0.99\textwidth]{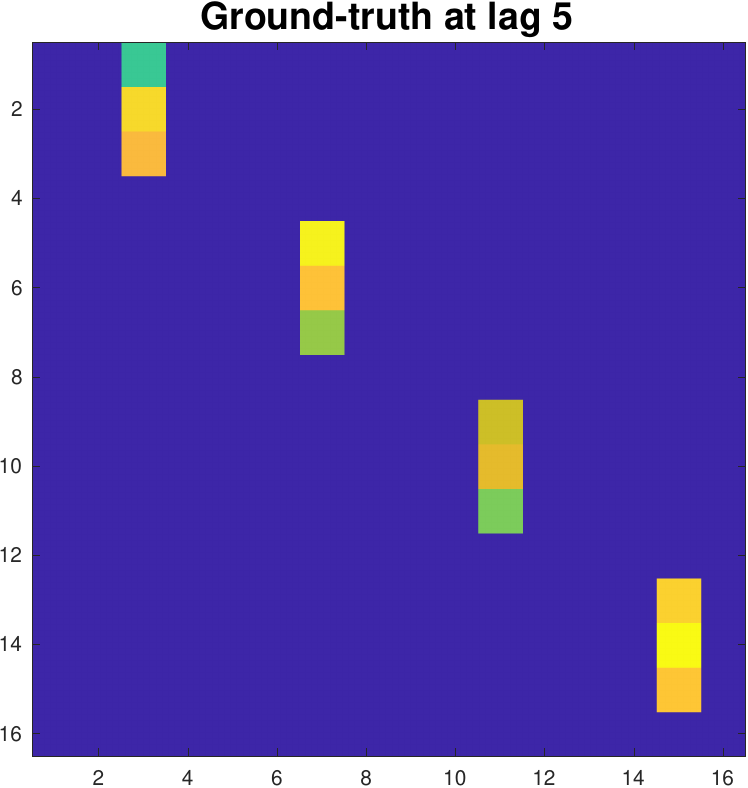}
         \end{subfigure} 
        \begin{subfigure}[b]{0.14\linewidth}
                 \centering
                 \includegraphics[width=0.99\textwidth]{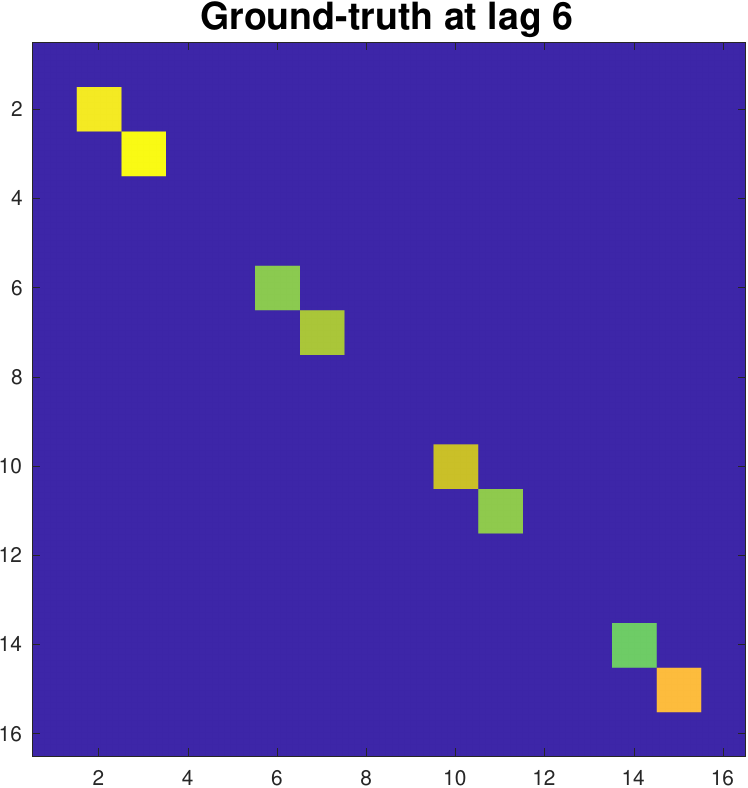}
         \end{subfigure}         
\\
         \begin{subfigure}[b]{0.14\linewidth}
                 \centering
                 \includegraphics[width=0.99\textwidth]{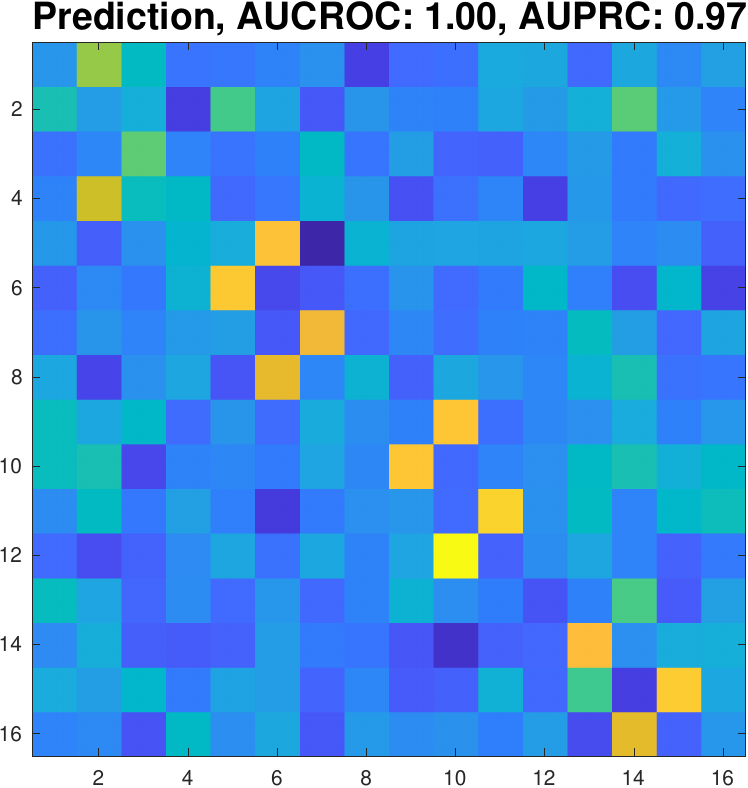}
         \end{subfigure}
         \begin{subfigure}[b]{0.14\linewidth}
                 \centering
                 \includegraphics[width=0.99\textwidth]{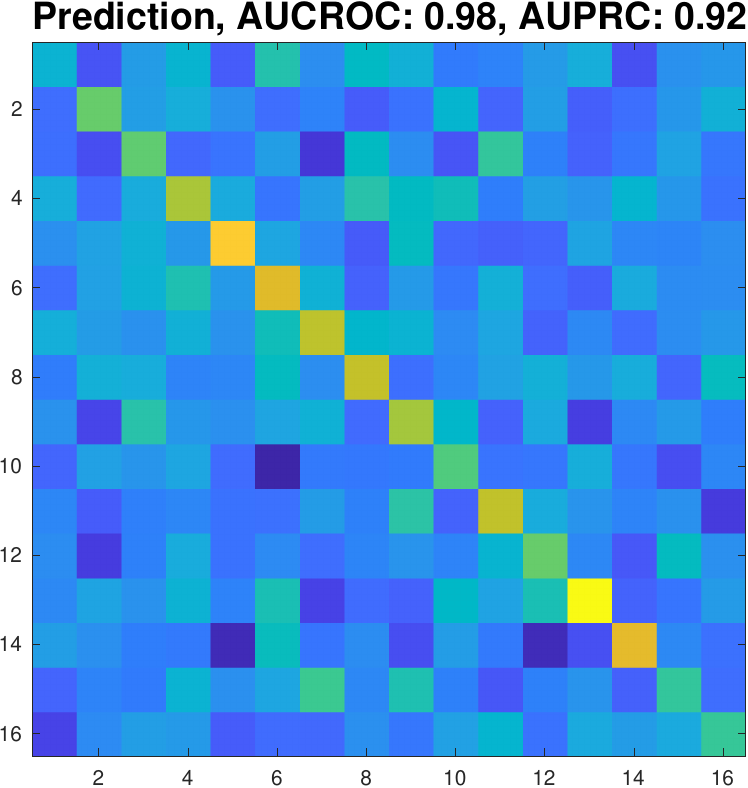}
         \end{subfigure} 
        \begin{subfigure}[b]{0.14\linewidth}
                 \centering
                 \includegraphics[width=0.99\textwidth]{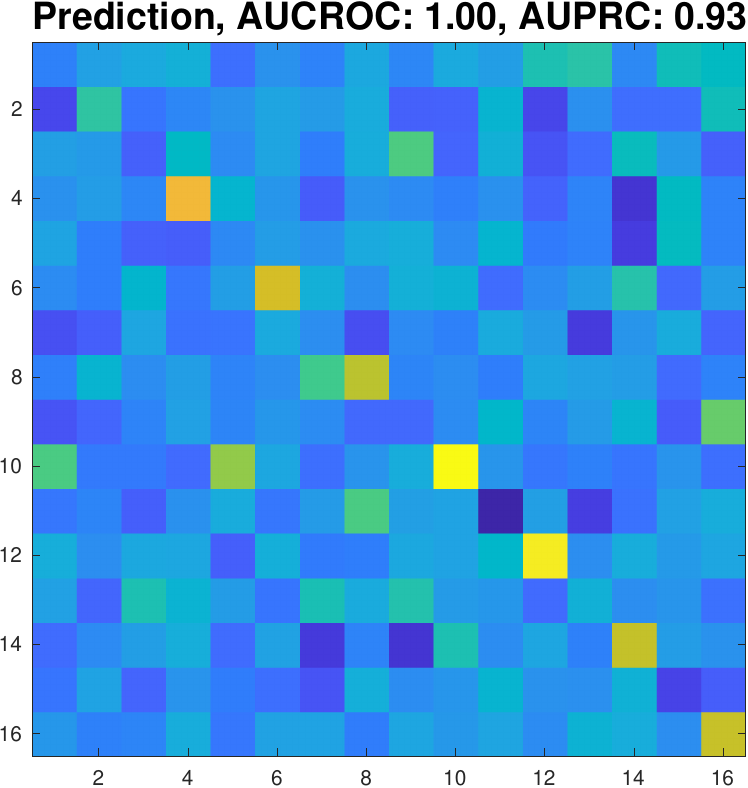}
         \end{subfigure}
                  \begin{subfigure}[b]{0.14\linewidth}
                 \centering
                 \includegraphics[width=0.99\textwidth]{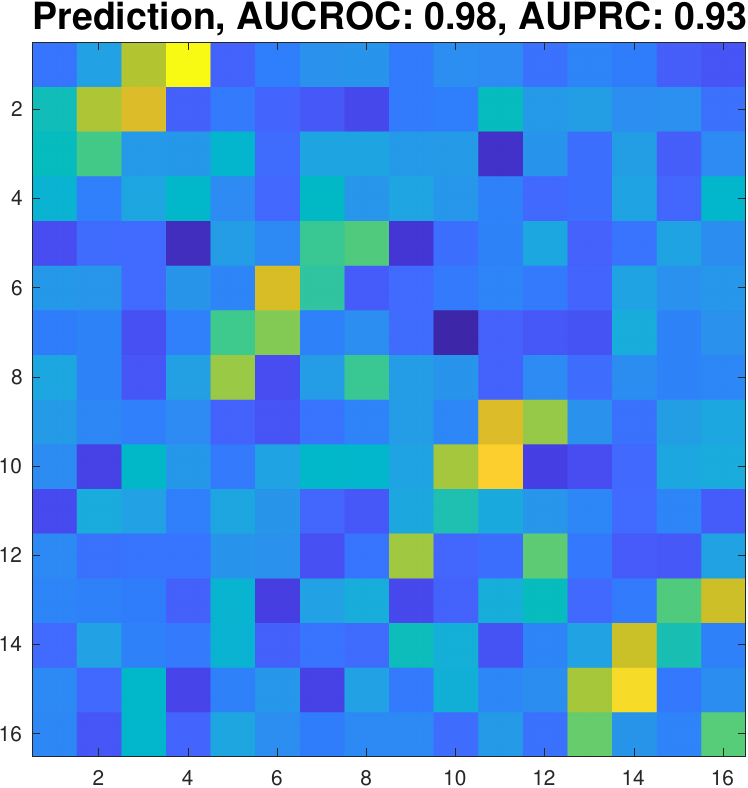}
         \end{subfigure}
         \begin{subfigure}[b]{0.14\linewidth}
                 \centering
                 \includegraphics[width=0.99\textwidth]{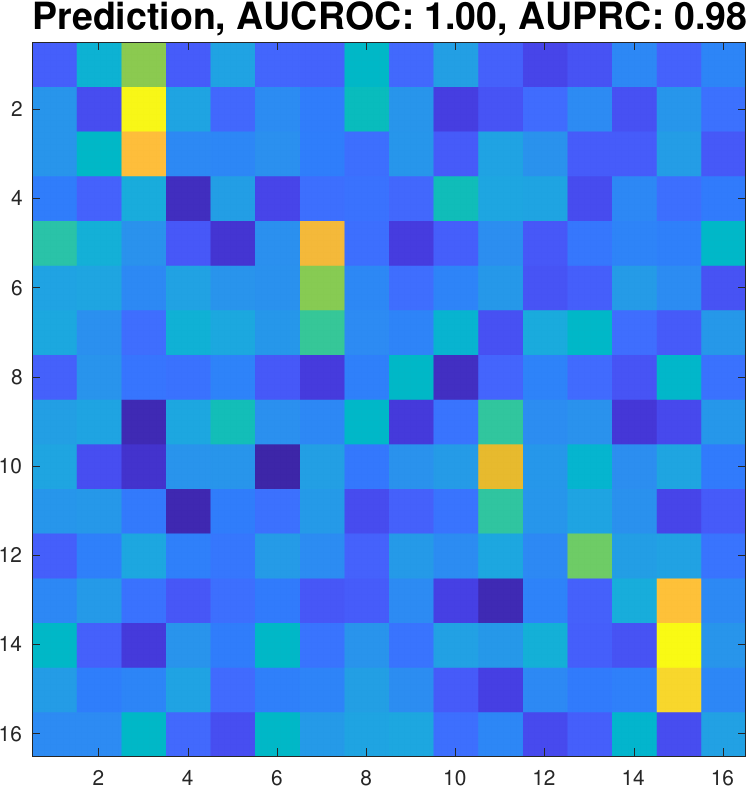}
         \end{subfigure} 
        \begin{subfigure}[b]{0.14\linewidth}
                 \centering
                 \includegraphics[width=0.99\textwidth]{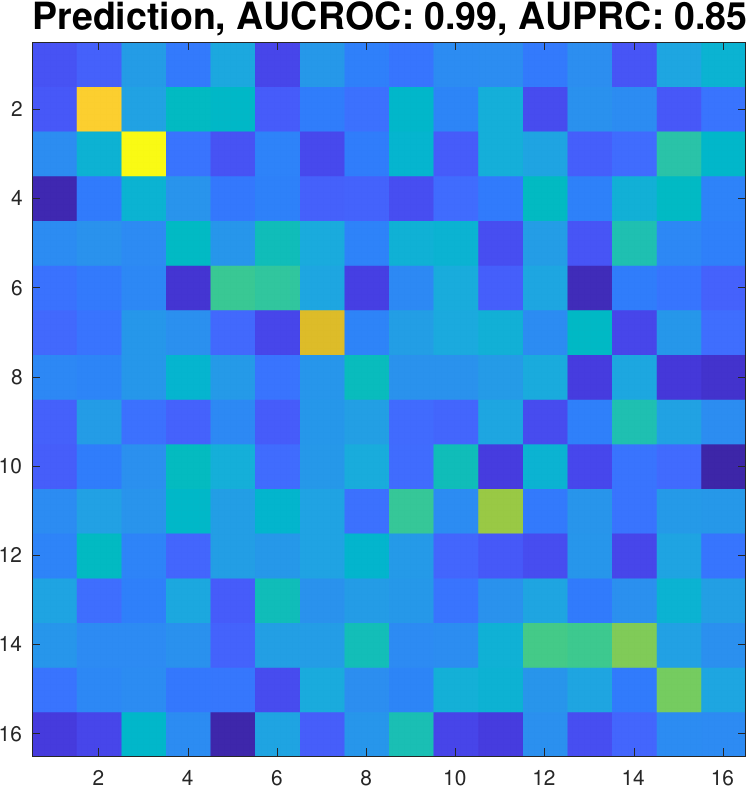}
         \end{subfigure}
\\
         \begin{subfigure}[b]{0.14\linewidth}
                 \centering
                 \includegraphics[width=0.99\textwidth]{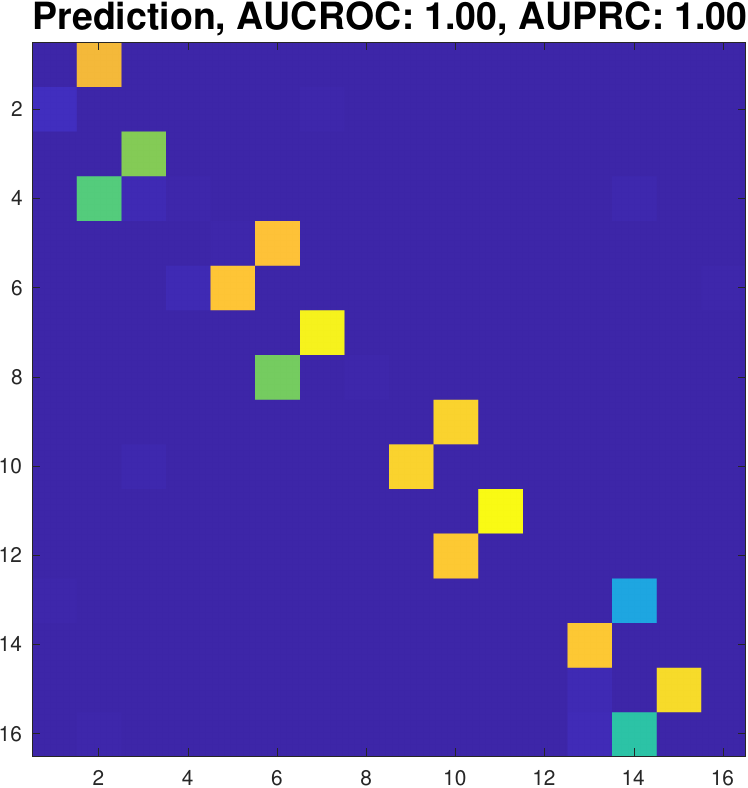}
         \end{subfigure}
         \begin{subfigure}[b]{0.14\linewidth}
                 \centering
                 \includegraphics[width=0.99\textwidth]{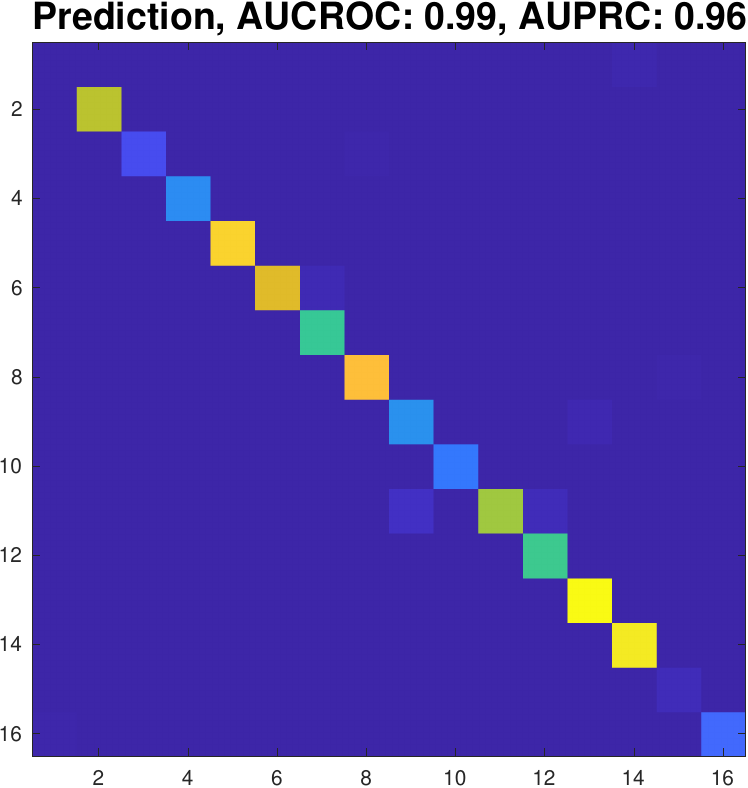}
         \end{subfigure} 
        \begin{subfigure}[b]{0.14\linewidth}
                 \centering
                 \includegraphics[width=0.99\textwidth]{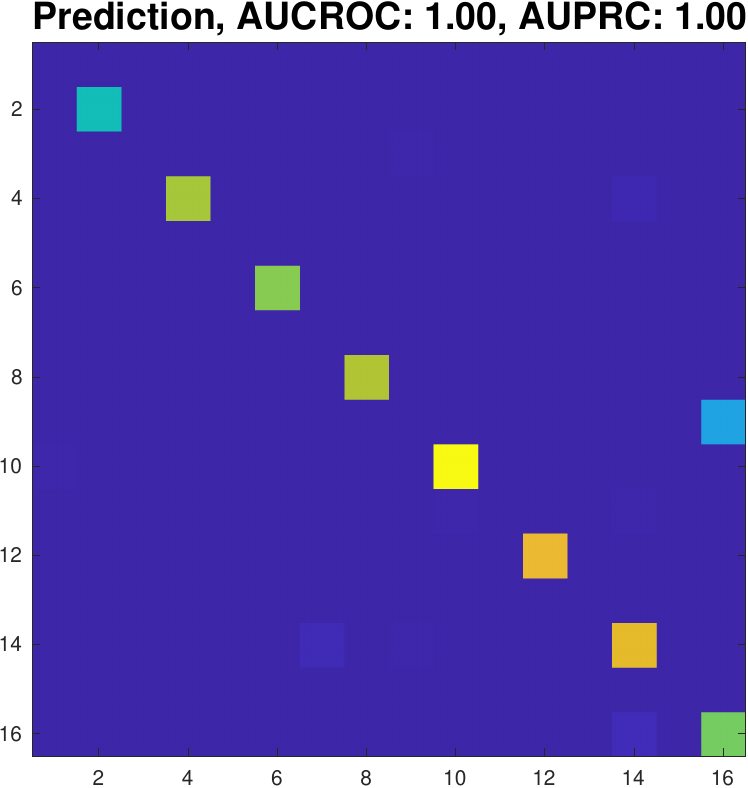}
         \end{subfigure}
           \begin{subfigure}[b]{0.14\linewidth}
                 \centering
                 \includegraphics[width=0.99\textwidth]{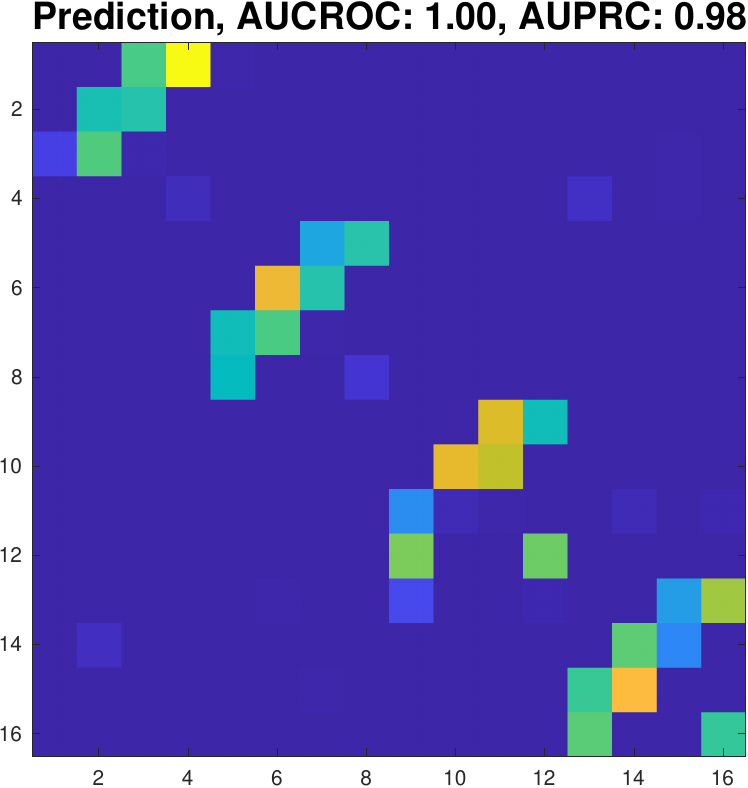}
         \end{subfigure}
         \begin{subfigure}[b]{0.14\linewidth}
                 \centering
                 \includegraphics[width=0.99\textwidth]{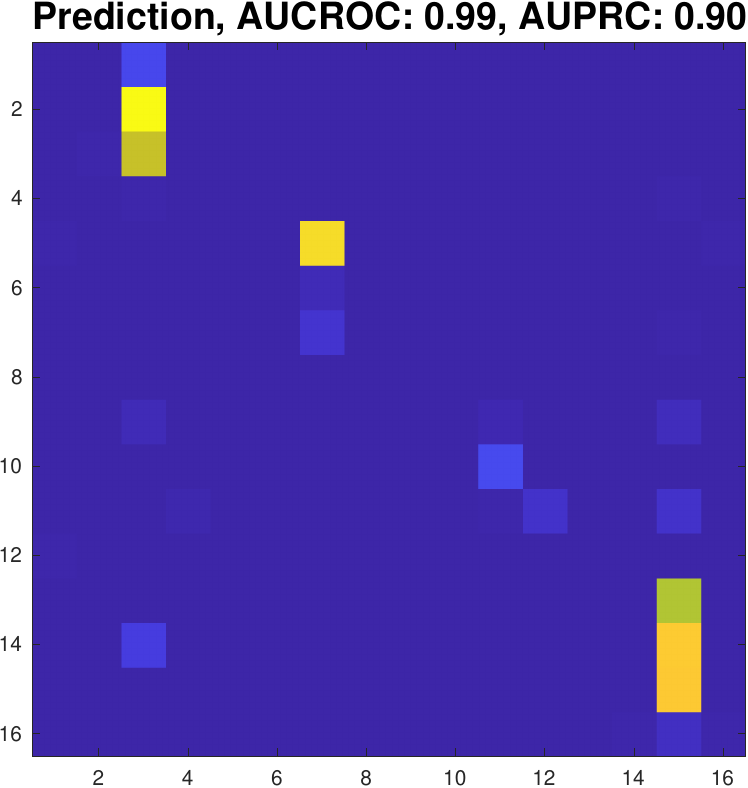}
         \end{subfigure} 
        \begin{subfigure}[b]{0.14\linewidth}
                 \centering
                 \includegraphics[width=0.99\textwidth]{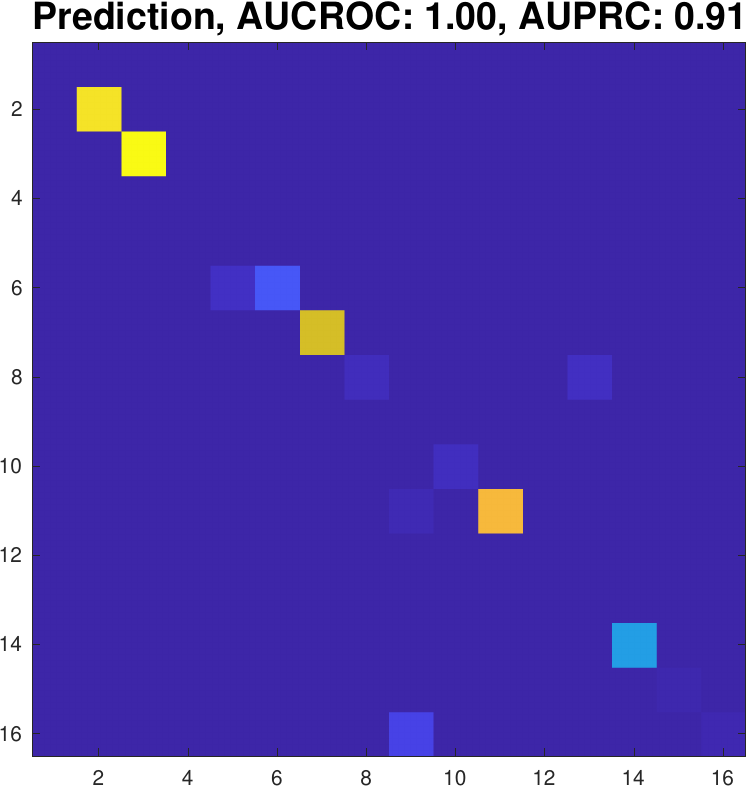}
         \end{subfigure}
\caption{Synthetic dataset. First to third rows: ground-truth graphs, coefficients of VAR, and GC graphs of ours at different lags. Columns: $\tau=1,\dots,6$. The AUCROC and AUPRC scores at each lag of VAR and ours are shown in the sub-captions. Mean AUCROC over all the lags is 0.99 (VAR) and 1.0 (ours); Mean AUPRC is 0.93 (VAR) and 0.96 (ours).}
\label{fig-syn-1}
\end{figure*}

\subsection{More Introduction to the Climate Reanalysis Data}
\label{sec-more-climate}
\begin{figure}[t]
\centering
\captionsetup[subfigure]{justification=centering}
        \centering
         \begin{subfigure}[b]{0.9\linewidth}
                 \centering
                 \includegraphics[width=0.99\textwidth]{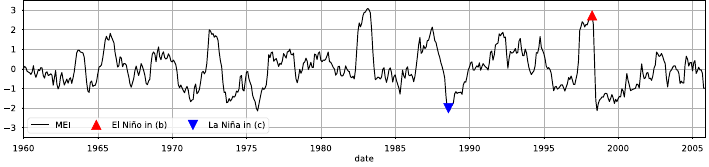}
                 \caption{}
         \end{subfigure}
         
         \begin{subfigure}[b]{0.45\linewidth}
                 \centering
                 \includegraphics[width=0.99\textwidth]{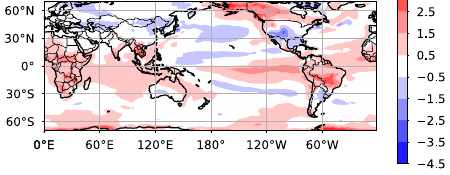}
                 \caption{}
         \end{subfigure}
          \begin{subfigure}[b]{0.45\linewidth}
                 \centering
                 \includegraphics[width=0.99\textwidth]{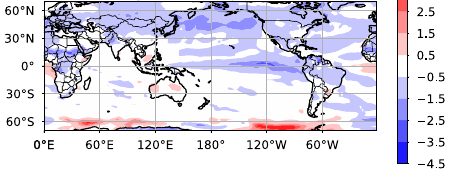}
                 \caption{}
         \end{subfigure}
\caption{(a) MEI index. (b) El Niño temperature anomaly in April 1998. (c) La Niña temperature anomaly August 1988.}
\label{fig-climate-vis-mei}
\end{figure}

The climate indices' names are shown in Table~\ref{tb-climate-index}.
Among the indices, the Multivariate El~Ni\~no Southern Oscillation Index (MEI) index is a representative example of the other climate indices in that the timeseries is associated with regionally distributed coherent responses in the atmosphere and surface ocean.
In Figure~\ref{fig-climate-vis-mei}(a), we plot the timeseries of MEI as an example, which characterises the El~Ni\~no / La~Ni\~na cycle. 
Positive values of the MEI are associated with El~Ni\~no periods, negative values are associated with La Ni\~na periods, with the magnitude of the index proportional to the strength of the event.
For example, according to the MEI index, there was a large El~Ni\~no in April of 1998.
Figure~\ref{fig-climate-vis-mei}(b) illustrates how much warmer (red) and cooler (blue) the surface air temperature was in that month with respect to the average April.
This map illustrates a large warm patch over the eastern Pacific ocean, which is typical of El~Ni\~no.
Conversely, the MEI indicates that August 1988 was a large La~Ni\~na, with an associated surface air temperature map illustrated in Figure~\ref{fig-climate-vis-mei}(c). This map illustrates how different the surface air temperature was with respect to an average August.
As is typical of La~Ni\~na events, the eastern Pacific region of anomalously cool.

\begin{figure*}[t]
\captionsetup[subfigure]{justification=centering}
        \centering
         \begin{subfigure}[b]{0.78\linewidth}
                 \centering
                 \includegraphics[width=0.99\textwidth]{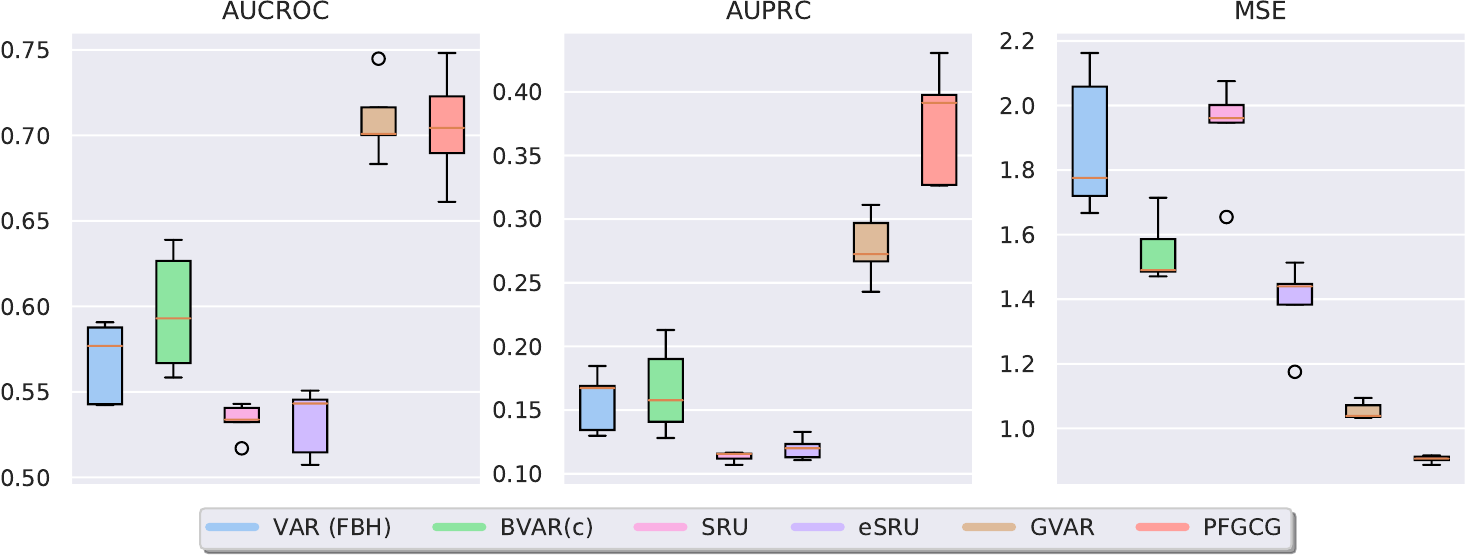}
         \end{subfigure}
         \begin{subfigure}[b]{0.85\linewidth}
                 \centering
                 \includegraphics[width=0.9\textwidth]{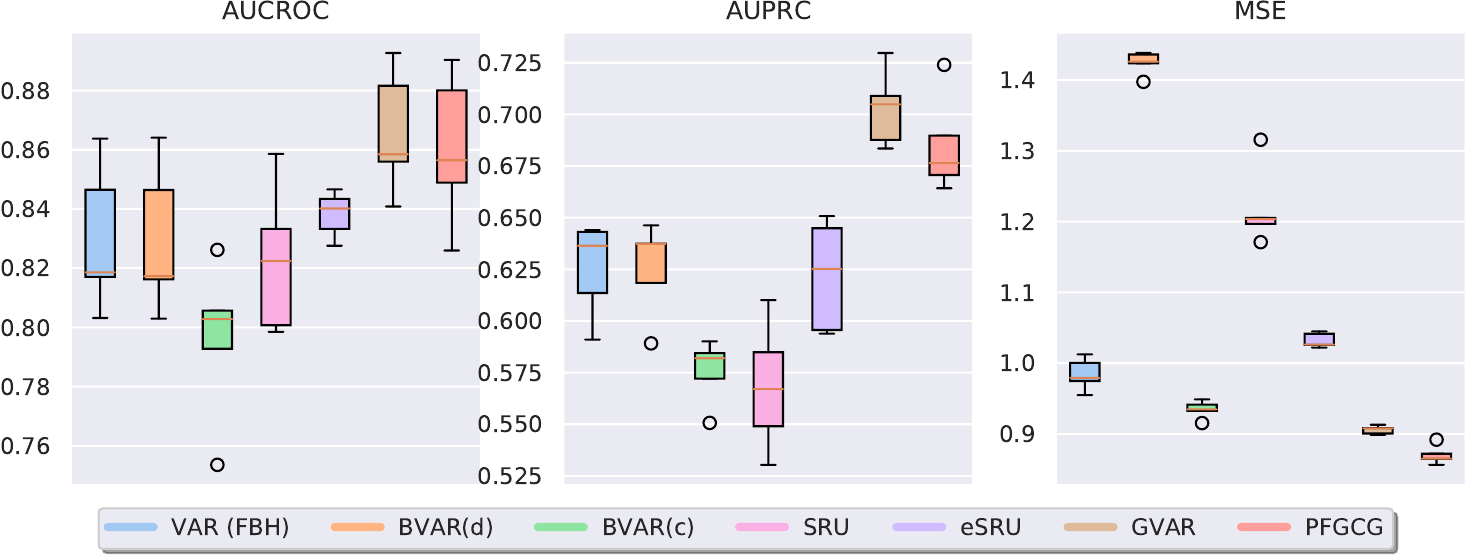}
         \end{subfigure} 
       
\caption{Lorenz 96 with $\taumax=1$. Upper: $T=100$, lower: $T=500$. VAR (FBH) and BVAR(d) failed to learn when $T=100$.}
\label{fig-l96-1}
\end{figure*}

\begin{figure*}[t]
\captionsetup[subfigure]{justification=centering}
        \centering
         \begin{subfigure}[b]{0.78\linewidth}
                 \centering
                 \includegraphics[width=0.99\textwidth]{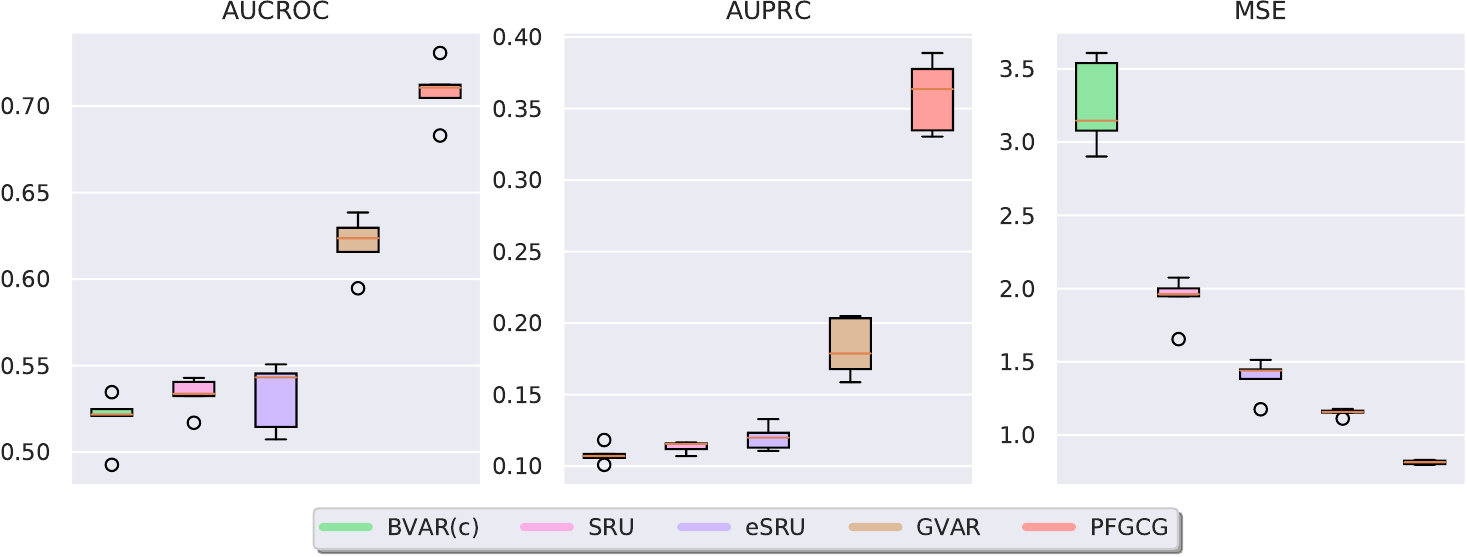}
         \end{subfigure}
         \begin{subfigure}[b]{0.85\linewidth}
                 \centering
                 \includegraphics[width=0.9\textwidth]{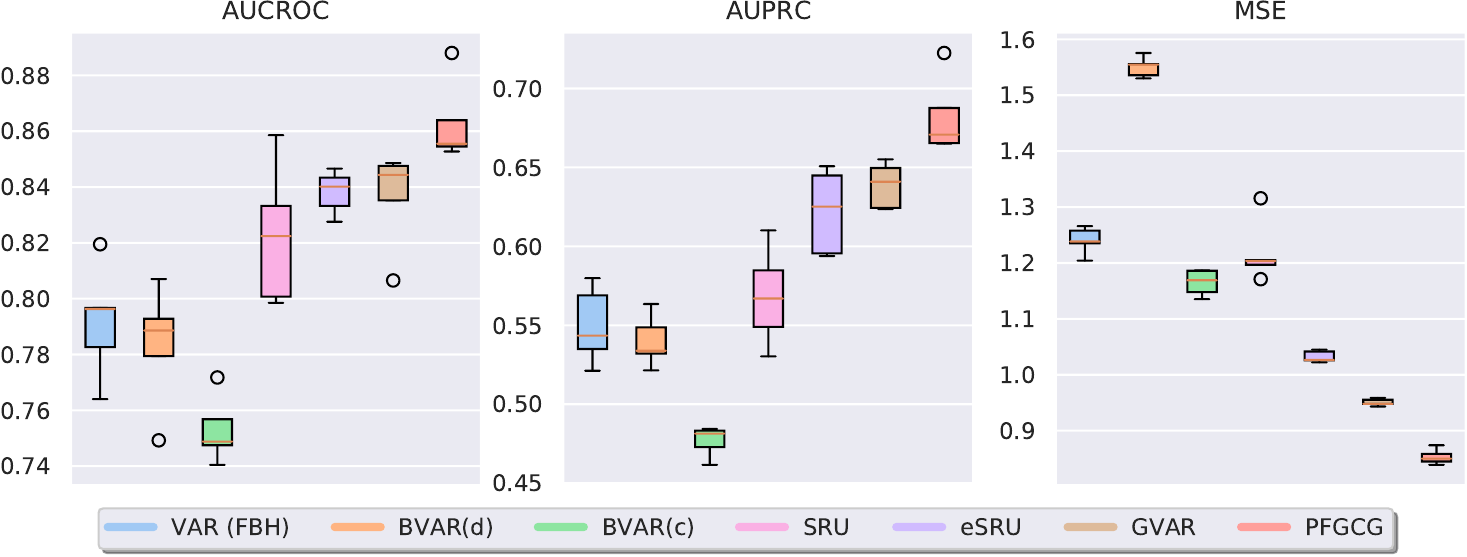}
         \end{subfigure} 
       
\caption{Lorenz 96 with $\taumax=3$. Upper: $T=100$, lower: $T=500$. BVAR(d) failed to learn when $T=100$.}
\label{fig-l96-3}
\end{figure*}

\begin{figure*}[t]
\captionsetup[subfigure]{justification=centering}
        \centering
         \begin{subfigure}[b]{0.78\linewidth}
                 \centering
                 \includegraphics[width=0.99\textwidth]{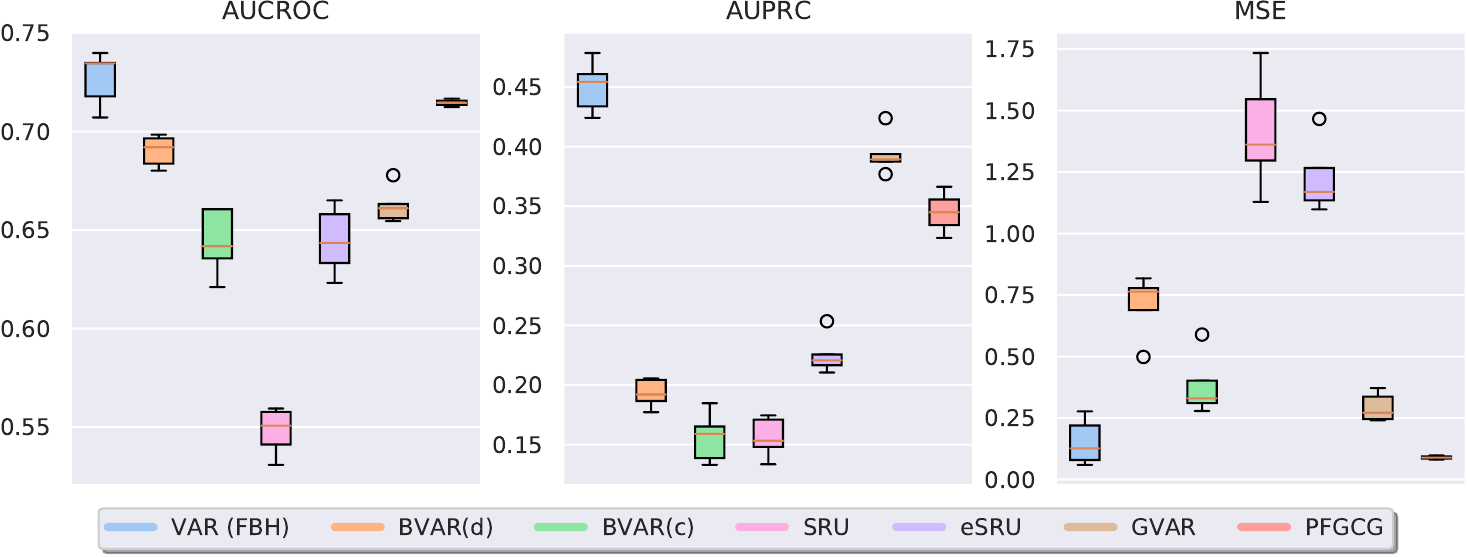}
         \end{subfigure}
         \begin{subfigure}[b]{0.85\linewidth}
                 \centering
                 \includegraphics[width=0.9\textwidth]{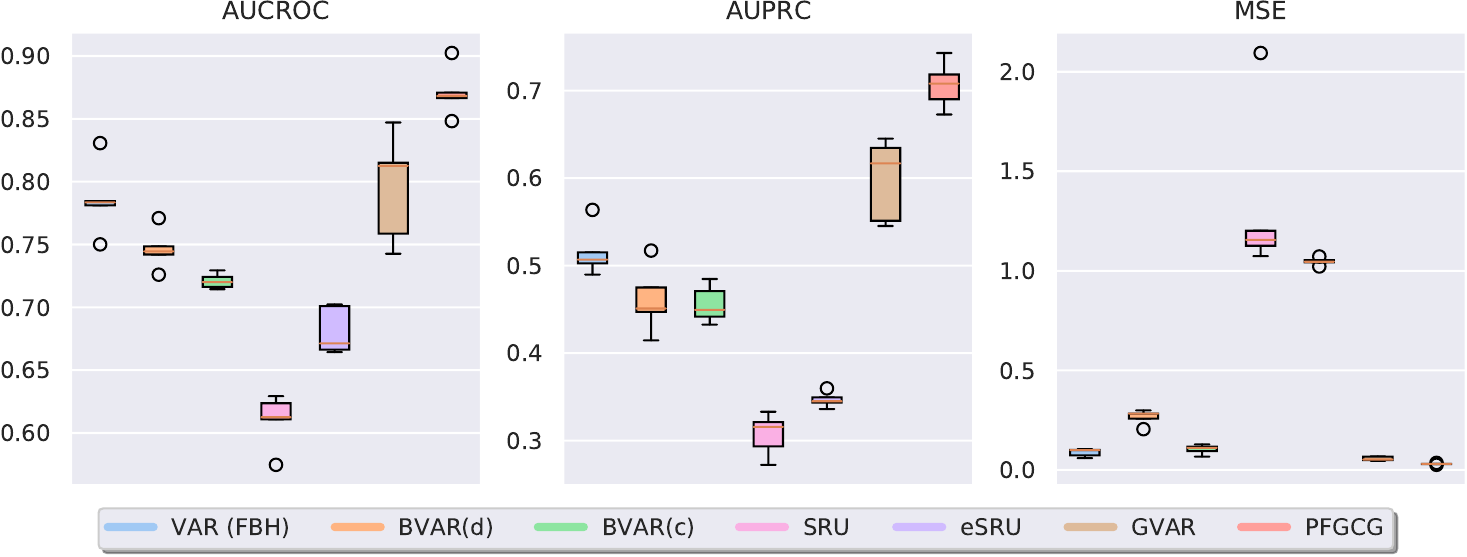}
         \end{subfigure} 
       
\caption{Lotka–Volterra with $\taumax=1$. Upper: $T=200$, lower: $T=500$.}
\label{fig-lv-1}
\end{figure*}

\begin{figure*}[t]
\captionsetup[subfigure]{justification=centering}
        \centering
         \begin{subfigure}[b]{0.78\linewidth}
                 \centering
                 \includegraphics[width=0.99\textwidth]{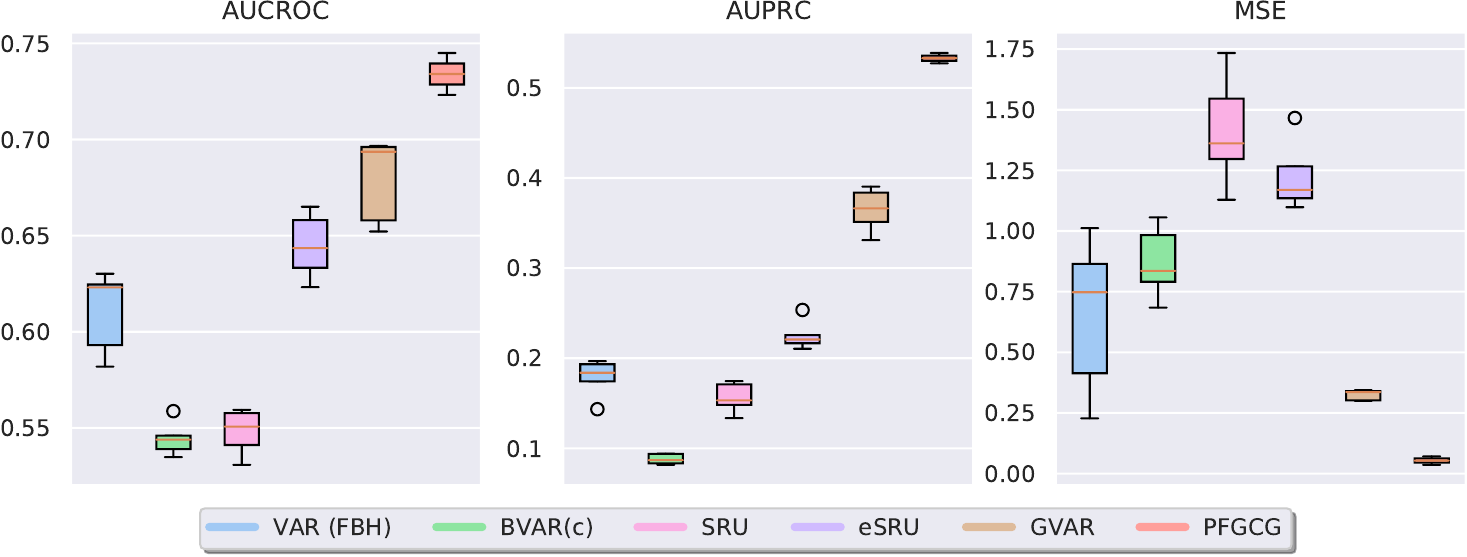}
         \end{subfigure}
         \begin{subfigure}[b]{0.85\linewidth}
                 \centering
                 \includegraphics[width=0.9\textwidth]{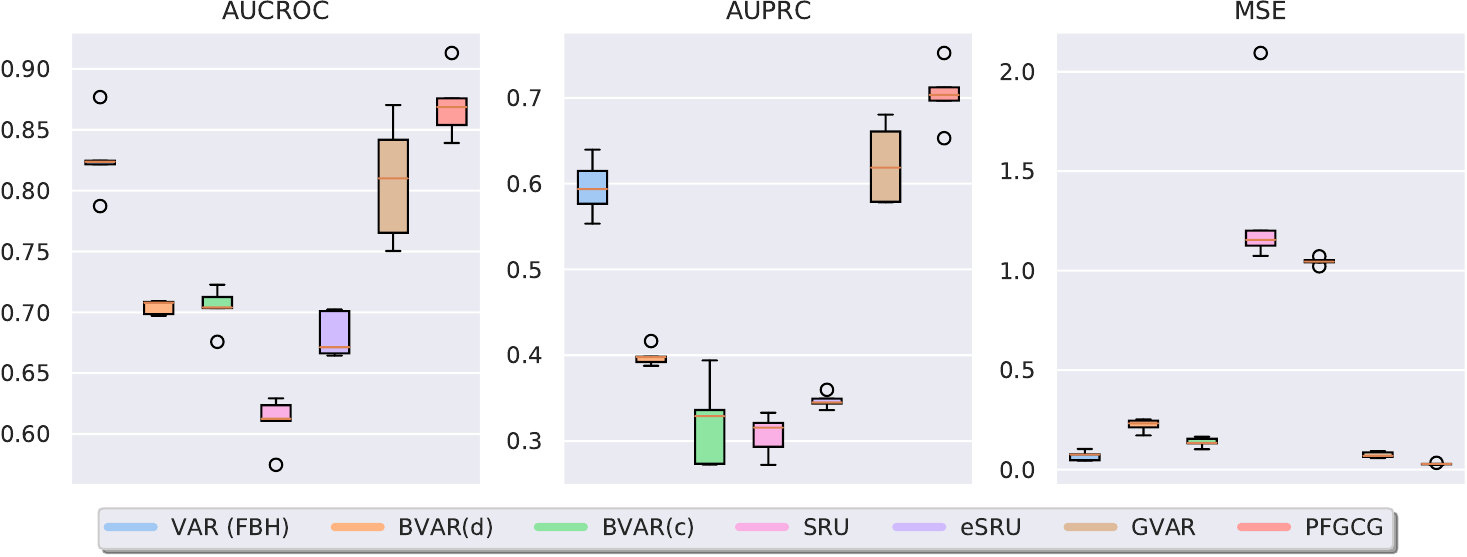}
         \end{subfigure} 
       
\caption{Lotka–Volterra with $\taumax=3$. Upper: $T=200$, lower: $T=500$. VAR (FBH) and BVAR(d) failed to learn when $T=200$.}
\label{fig-lv-3}
\end{figure*}

\begin{figure*}[t]
\captionsetup[subfigure]{justification=centering}
        \centering
         \begin{subfigure}[b]{0.78\linewidth}
                 \centering
                 \includegraphics[width=0.99\textwidth]{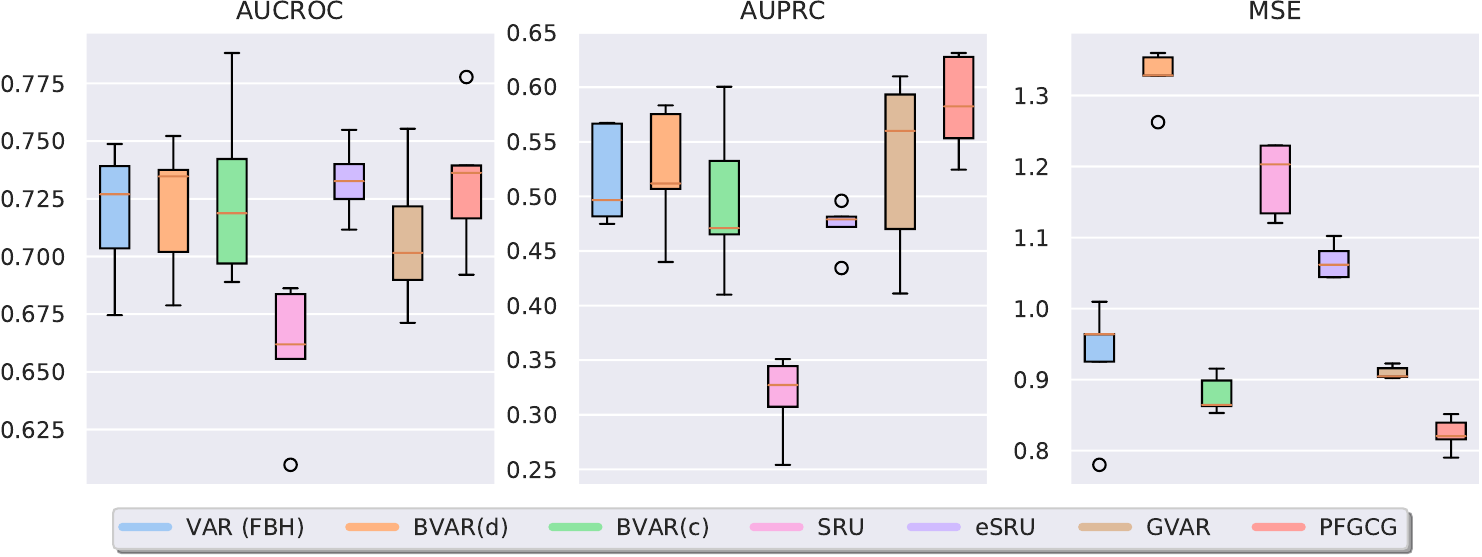}
         \end{subfigure}
\caption{FMRI  with $\taumax=1$.}
\label{fig-fmri-1}
\end{figure*}

\begin{figure*}[t]
\captionsetup[subfigure]{justification=centering}
        \centering
         \begin{subfigure}[b]{0.78\linewidth}
                 \centering
                 \includegraphics[width=0.99\textwidth]{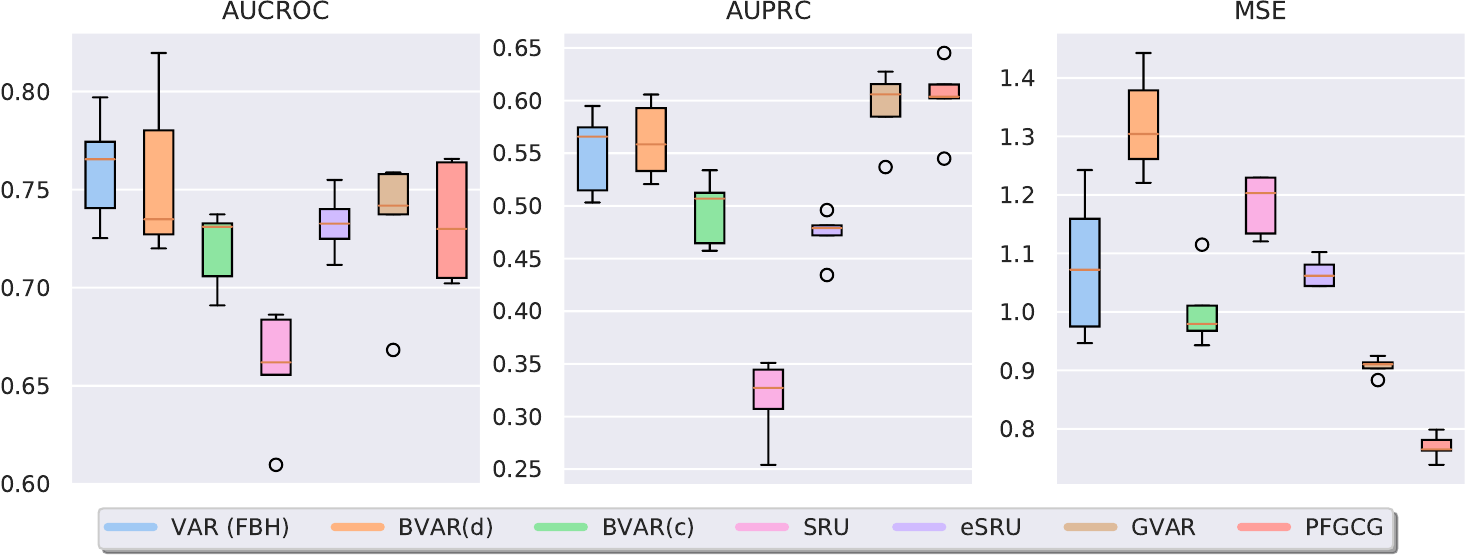}
         \end{subfigure}
\caption{FMRI  with $\taumax=3$.}
\label{fig-fmri-3}
\end{figure*}

\begin{table*}[]
\rowa{1.3}
\centering
\caption{SHD for VAR (FBH), GVAR, and PFGCG with $\taumax=1$. Means and standard derivations are computed over 5 replicates on each dataset.}
\label{tb-shd-1}
\begin{tabular}{@{}ccccccc@{}} \toprule

          & \multicolumn{2}{c}{L96}                                  & \multicolumn{2}{c}{LV}                                  & FMRI            \\ 
          & \multicolumn{1}{c}{$T=100$}           & $T=500$          & \multicolumn{1}{c}{$T=200$}           & $T=500$         &                 \\ \midrule
VAR (FBH) & \multicolumn{1}{c}{-}                 & 72.00$\pm$4.69   & \multicolumn{1}{c}{125.60$\pm$21.88}                 & 451.40$\pm$69.85 & 26.00$\pm$1.10  \\ 
GVAR      & \multicolumn{1}{c}{373.20$\pm$39.77} & 93.60$\pm$55.23 & \multicolumn{1}{c}{147.80$\pm$80.36} & 111.00$\pm$39.75 & 51.00$\pm$16.79 \\ 
PFGCG     & \multicolumn{1}{c}{112.91$\pm$1.95}   & 71.43$\pm$3.52   & \multicolumn{1}{c}{186.13$\pm$21.84}    & 49.73$\pm$4.72  & 25.17$\pm$1.43  \\ \bottomrule
\end{tabular}
\end{table*}

\begin{table*}[]
\rowa{1.3}
\centering
\caption{SHD for VAR (FBH), GVAR, and PFGCG with $\taumax=3$. Means and standard derivations are computed over 5 replicates on each dataset.}
\label{tb-shd-3}
\begin{tabular}{@{}ccccccc@{}} \toprule

          & \multicolumn{2}{c}{L96}                                  & \multicolumn{2}{c}{LV}                                  & FMRI            \\ 
          & \multicolumn{1}{c}{$T=100$}           & $T=500$          & \multicolumn{1}{c}{$T=200$}           & $T=500$         &                 \\ \midrule
VAR (FBH) & \multicolumn{1}{c}{-}                 & 79.60$\pm$3.32   & \multicolumn{1}{c}{84.40$\pm$4.18}                 & 75.40$\pm$5.85 & 25.80$\pm$1.33  \\ 
GVAR      & \multicolumn{1}{c}{577.40$\pm$47.51} & 133.20$\pm$107.22 & \multicolumn{1}{c}{213.80$\pm$117.98} & 128.60$\pm$68.31 & 50.00$\pm$18.99 \\ 
PFGCG     & \multicolumn{1}{c}{114.75$\pm$1.68}   & 73.18$\pm$2.67   & \multicolumn{1}{c}{75.30$\pm$9.21}    & 45.50$\pm$2.34  & 24.78$\pm$1.08  \\ \bottomrule
\end{tabular}
\end{table*}

\begin{table}[]
\rowa{1.3}
\centering
\caption{Climate index names}
\label{tb-climate-index}
\begin{tabular}{@{}ccc@{}}
\toprule
Index name                                                & Abbreviation \\ \midrule
Atlantic Oscillation                                      & AO           \\
Indian Ocean Dipole                                       & IOD          \\
Multivariate El Niño Southern Oscillation Index          & MEI          \\
North Atlantic Oscillation (positive and negative phases) & NAO+/-       \\
Atlantic Ridge patterns                                   & AR           \\
Scandinavian blocking patterns                            & SCAND        \\
Pacific North American patterns                           & PNA          \\
Pacific South American patterns                           & PSA1/2       \\
Southern Annular Mode                                     & SAM          \\
Wheeler-Hendon Madden-Julian oscillation                  & RMM1/2\\
\bottomrule
\end{tabular}
\end{table}

\end{document}